\documentclass[10pt,twocolumn,letterpaper]{article}

\usepackage{iccv}
\usepackage{times}
\usepackage{epsfig}
\usepackage{graphicx}
\usepackage{amsmath}
\usepackage{amssymb}

\usepackage[pagebackref=true,breaklinks=true,letterpaper=true,colorlinks,bookmarks=false]{hyperref}
\iccvfinalcopy 

\def\iccvPaperID{9236} 

\ificcvfinal\fi

\usepackage{bm}
\usepackage{textcomp} 

\usepackage{multirow}
\usepackage{subcaption}
\usepackage{pifont} 
\usepackage{makecell} 
\usepackage{tablefootnote}
\usepackage[ruled,vlined]{algorithm2e}

\SetCommentSty{mycommfont}
\usepackage{dsfont} 

\usepackage{etoolbox} %

\makeatletter
\@namedef{ver@everyshi.sty}{}
\makeatother
\usepackage{tikz}
\usetikzlibrary{arrows,calc,decorations.markings,math,arrows.meta,decorations.pathreplacing}
\definecolor{mediumtealblue}{rgb}{0.0, 0.33, 0.71}
\definecolor{darkpastelgreen}{rgb}{0.01, 0.75, 0.24}
\definecolor{azure(colorwheel)}{rgb}{0.0, 0.5, 1.0}

\newcommand{\D}{\mathsf{D}}
\newcommand{\G}{\mathsf{G}}
\newcommand{\F}{\mathsf{F}}

\newcommand{\ours}{\textcolor{azure(colorwheel)}{(ours)}}

\newcommand{\lhqsize}{90k}

\newcommand{\R}{\mathbb{R}}

\newcommand{\apref}[1]{\ref{#1}}

\begin{document}

\title{Aligning Latent and Image Spaces to Connect the Unconnectable}

\author{Ivan Skorokhodov\\
KAUST\\
{\tt\small iskorokhodov@gmail.com}
\and
Grigorii Sotnikov\\
Gradient, HSE, Skoltech\\
{\tt\small gdsotnikov@edu.hse.ru}
\and
Mohamed Elhoseiny\\
KAUST\\
{\tt\small mohamed.elhoseiny@kaust.edu.sa}
}

\makeatletter
\let\@oldmaketitle\@maketitle
\renewcommand{\@maketitle}{\@oldmaketitle
\myfigure\bigskip}
\makeatother
\newcommand\myfigure{%
  \makebox[0pt]{\hspace{17.5cm}\includegraphics[width=\linewidth]{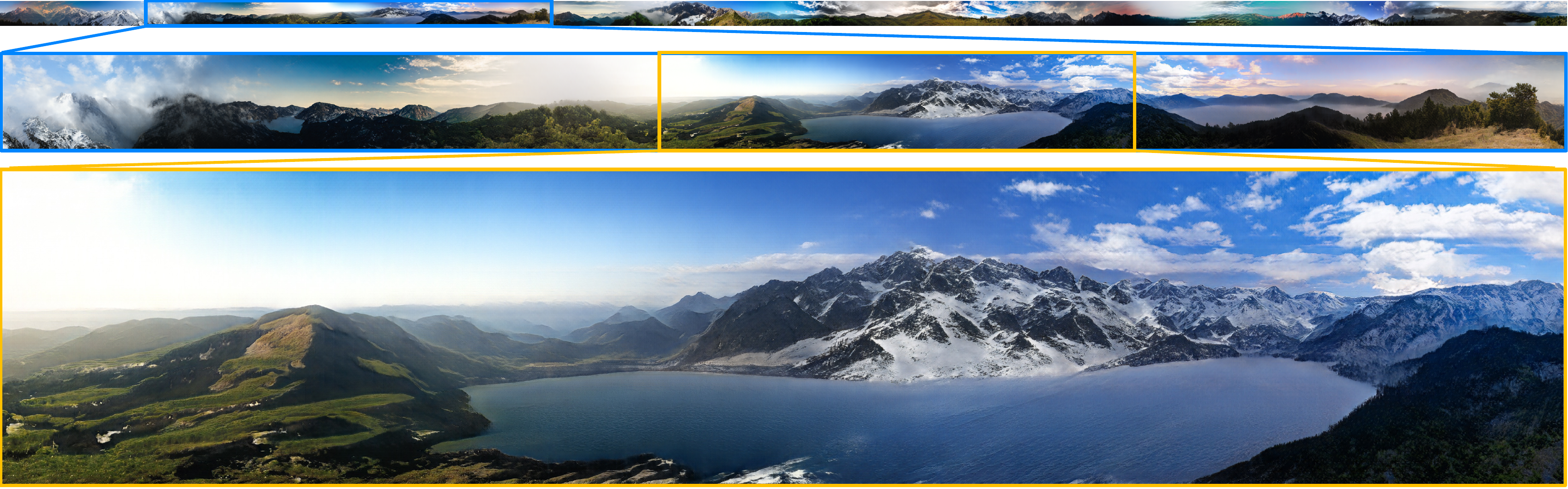}}
  \\
  \refstepcounter{figure}\normalfont{Figure~\thefigure: Our method can generate infinite images of diverse and complex scenes that transition naturally from one into another. It does so without any conditioning and trains without any supervision from a dataset of unrelated square images.}
  \label{fig:teaser}
}

\maketitle
\ificcvfinal\thispagestyle{empty}\fi

\begin{abstract}
In this work, we develop a method to generate infinite high-resolution images with diverse and complex content.
It is based on a perfectly equivariant generator with synchronous interpolations in the image and latent spaces.
Latent codes, when sampled, are positioned on the coordinate grid, and each pixel is computed from an interpolation of the nearby style codes.
We modify the AdaIN mechanism to work in such a setup and train the generator in an adversarial setting to produce images positioned between any two latent vectors.
At test time, this allows for generating complex and diverse infinite images and connecting any two unrelated scenes into a single arbitrarily large panorama.
Apart from that, we introduce LHQ: a new dataset of \lhqsize high-resolution nature landscapes.
We test the approach on LHQ, LSUN Tower and LSUN Bridge and outperform the baselines by at least 4 times in terms of quality and diversity of the produced infinite images.
The project website is located at \href{https://universome.github.io/alis}{https://universome.github.io/alis}.
\end{abstract}

\section{Introduction}

Modern image generators are typically designed to synthesize pictures of some fixed size and aspect ratio.
The real world, however, continues outside the boundaries of any captured photograph, and so to match this behavior, several recent works develop architectures to produce infinitely large images \cite{InfinityGAN, TamingTransformers, LocoGAN}, or images that partially extrapolate outside their boundary \cite{CocoGAN, INR_GAN, LocoGAN}.

Most of the prior work on infinite image generation focused on the synthesis of homogeneous texture-like patterns \cite{SpatialGAN, PS_GAN, InfinityGAN} and did not explore the infinite generation of complex scenes, like nature or city landscapes.
The critical challenge of generating such images compared to texture synthesis is making the produced frames globally consistent with one another: when a scene spans across several frames, they should all be conditioned on some shared information.
To our knowledge, the existing works explored three ways to achieve this: 1) fit a separate model per scene, so the shared information is encoded in the model's weights (e.g., \cite{InfinityGAN, SinGAN, PosEncSinGAN, TileGAN}); 2) condition the whole generation process on a global latent vector \cite{LocoGAN, CocoGAN, INR_GAN}; and 3) predict spatial latent codes autoregressively \cite{TamingTransformers}.

The first approach requires having a large-resolution photograph (like satellite images) and produces pictures whose variations in style and semantics are limited to the given imagery.
The second solution can only perform some limited extrapolation since using a single global latent code cannot encompass the diversity of an infinite scenery (as also confirmed by our experiments).
The third approach is the most recent and principled one, but the autoregressive inference is dramatically slow \cite{NonAutoregressiveNMT}: generating a single $256^2$ image with \cite{TamingTransformers}'s method takes us {$\sim$}10 seconds on a single V100 GPU.

This work, like \cite{TamingTransformers}, also seeks to build a model with global consistency and diversity in its generation process.
However, in contrast to \cite{TamingTransformers}, we attack the problem from a different angle.
Instead of slowly generating \textit{local} latent codes autoregressively (to make them coherent with one another), we produce several \textit{global} latent codes independently and train the generator to connect them.

\begin{figure}
    \centering
    \includegraphics[width=\linewidth]{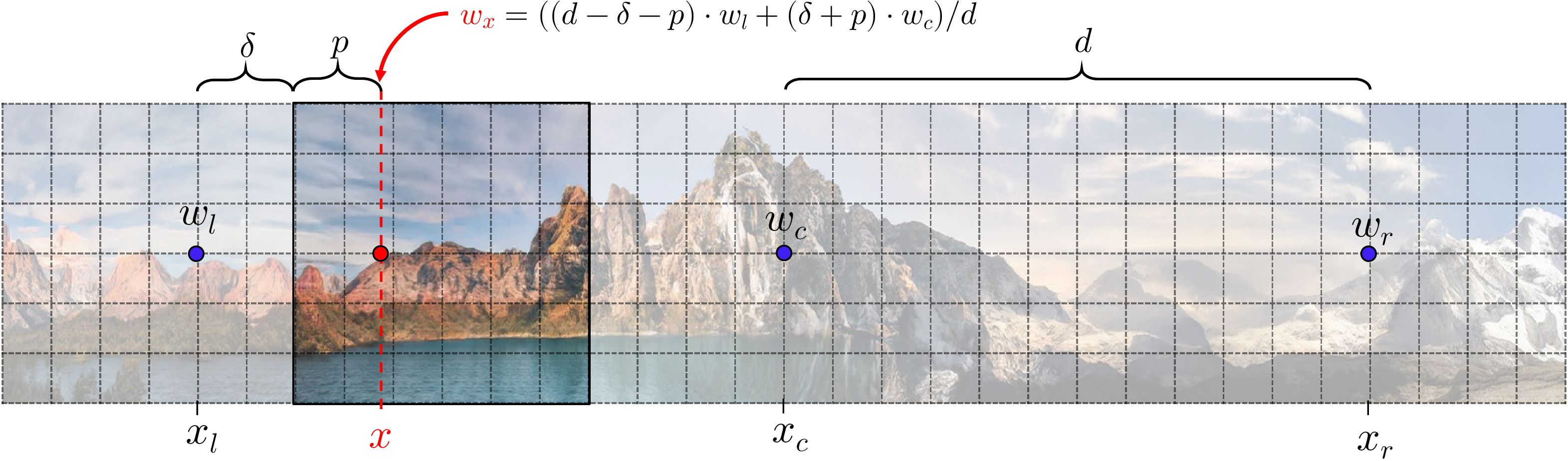}
    \caption{Illustration of our alignment procedure. We position initial latent codes (\textit{anchors}) on the 2D coordinate space and compute latent code $\bm w_x$ for each position $x$ as a linear interpolation between its two neighboring anchors.}
    \label{fig:alignment}
\end{figure}

We build on top of the recently proposed coordinate-based decoders that produce images based on pixels' coordinate locations \cite{CoordConv, CocoGAN, INR_GAN, CIPS, LIIF} and develop the above idea in the following way.
Latent codes, when sampled, are positioned on the 2D coordinate space --- they constitute the ``anchors'' of the generation process.
Next, each image patch is computed independently from the rest of the image, and the latent vector used for its generation is produced by linear interpolation between nearby anchors.
If the distance $d \in \R$ between the anchors is sufficiently large, they cover large regions of space with the common global context and make the neighboring frames in these regions be semantically coherent.
This idea of aligning latent and image spaces (ALIS) is illustrated in Figure~\ref{fig:alignment}.

Our model is GAN-based \cite{GAN} and the generator is trained to produce plausible images from any position in space.
This is in high contrast to existing coordinate-based approaches that generate samples only from $[0, 1]^2$ coordinates area, which constitutes a single frame size.
To make the model generalize to any position in space, we do not input global coordinates information.
Instead, we input only its position relative to the neighboring anchors, which, in turn, could be located arbitrarily on the $x$-axis.

We utilize StyleGAN2 architecture \cite{StyleGAN2} for our method and modify only its generator component.
Originally, StyleGAN2 passes latent vectors into the decoder by modulating and demodulating convolutional weights, an adjustment of adaptive instance normalization layer (AdaIN) \cite{AdaIN}.
For our generator, we adjust AdaIN differently to make it work with the coordinate-based latent vectors and develop Spatially-Aligned AdaIN (SA-AdaIN), described in Sec~\ref{sec:method}.
Our model is trained in a completely unsupervised way from a dataset of unrelated square image crops, i.e. it never sees full panorama images or even different parts of the same panorama during training.
By training it to produce realistic images located between arbitrary anchors describing different content (for example, mountains and a forest), it learns to connect unrelated scenes into a single panorama.
This task can be seen as learning to generate camera transitions between two viewpoints located at semantically very different locations.

We test our approach on several LSUN categories and Landscapes HQ (LHQ): a new dataset consisting of 90k high-resolution nature landscape images that we introduce in this work.
We outperform the existing baselines for all the datasets in terms of infinite image quality by at least 4 times and at least 30\% in generation speed.


\section{Related work}

\textbf{GANs}.
During the past few years, GAN-based models achieved photo-realistic quality in 2D image generation \cite{GAN, BigGAN, StyleGAN2}.
Much work has been done to design better optimization objectives \cite{WGAN, WGAN_GP, GeometricGAN} and regularization schemes \cite{SpectralNorm, R1_reg, NumericsOfGANs, ConsistencyReg_GAN}, make the training more compute \cite{NotSoBigGAN, SWAGAN, LightweightGAN} and data \cite{SGA, DiffAug} efficient and develop robust evaluation metrics \cite{TTUR, ImprovedTechniquesGANs, Hype, PrecisionRecallGAN}.
Compared to VAEs \cite{VAE}, GANs do not provide a ready-to-use encoder.
Thus much work has been devoted to either training a separate encoder component \cite{ALAE, SwappingAE} or designing procedures of embedding images into the generator's latent space \cite{Image2StyleGAN, PULSE}.

\textbf{Coordinates conditioning}.
Coordinates conditioning is the most popular among the NeRF-based \cite{NeRF, NerfInTheWild} and occupancy-modeling \cite{OccupancyNetworks, IM-NET, DeepMetaFunctionals} methods.
\cite{GIRAFFE, pi-GAN, Graf} trained a coordinate-based generator that models a volume which is then rendered and passed to a discriminator.
However, several recent works demonstrated that providing positional information can help the 2D world as well.
For example, it can improve the performance on several benchmarks \cite{CoordConv, CIPS} and lead to the emergence of some attractive properties like extrapolation \cite{CocoGAN, INR_GAN} or super-resolution \cite{INR_GAN, LIIF}. 
An important question in designing coordinate-based methods is how to embed positional information into a model \cite{Transformer, Conv_seq2seq}.
Most of the works rely either on raw coordinates \cite{CoordConv, LocoGAN} or periodic embeddings with log-linearly distributed frequencies \cite{NeRF}.
\cite{SIREN, FourierINR} recently showed that using Gaussian distributed frequencies is a more principled approach.
\cite{Nerfies} developed a progressive growing technique for positional embeddings.

\textbf{Infinite image generation}.
Existing works on infinite image generation mainly consider the generation of only texture-like and pattern-like images \cite{SpatialGAN, PS_GAN, TileGAN, InfinityGAN, GramGAN}, making it similar to procedural generation \cite{ProcGenOfCities, ProcGenOfBuildings}.
SinGAN \cite{SinGAN} learn a GAN model from a single image and is able to produce its (potentially unbounded) variations.
Generating infinite images with diverse, complex and globally coherent content is a more intricate task since one needs to seamlessly stitch both local and global features.
Several recent works advanced on this problem by producing images with non-fixed aspect ratios \cite{LocoGAN, TamingTransformers} and could be employed for the infinite generation.
LocoGAN~\cite{LocoGAN} used a single global latent code to produce an image to make the generation globally coherent, but using a \textit{single} global code leads to content saturation (as we show in Sec~\ref{sec:experiments}).
Taming Transformers (TT)~\cite{TamingTransformers} proposed to generate global codes autoregressively with Transformer \cite{Transformer}, but since their decoder uses GroupNorm \cite{GroupNorm} under the hood, the infinite generation is constrained by the GPU memory at one's disposal (see Appendix~\apref{ap:taming-transformers} for the details).
\cite{InfiniteNature} proposes both a model and a dataset to generate from a single RGB frame a sequence of images along a camera trajectory.
Our approach shares some resemblance to image stitching \cite{Image_stitching} but intends to generate \textit{an entire scene} that lies between the given two images instead of concatenating two existing pictures into a single one without seams.
\cite{InfiniteImages} constructed an infinite image by performing image retrieval + stitching from a vast image collection.

\textbf{Image extrapolation}.
Another close line of research is image extrapolation (or image ``outpainting'' \cite{ImageOutpainting, GAN_based_outpainting}), i.e., predicting the surrounding context of an image given only its part.
The latest approaches in this field rely on using GANs to predict an outlying image patch \cite{SpiralGAN, Boundless, WideContextExtrapolation}.
The fundamental difference of these methods compared to our problem design is the reliance on an input image as a starting point of the generation process.

\textbf{Adaptive Normalization}.
Instance-based normalization techniques were initially developed in the style transfer literature \cite{GatysStyleTransfer}.
Instance Normalization \cite{InstanceNorm} was proposed to improve feed-forward style transfer \cite{ImprovedTextureNetworks} by replacing content image statistics with the style image ones.
CIN \cite{Conditional_IN} learned separate scaling and shifting parameters for each style.
AdaIN \cite{AdaIN} developed the idea further and used shift and scaling values produced by a separate module to perform style transfer from an arbitrary image.
Similar to StyleGAN \cite{StyleGAN}, our architecture uses AdaIN \cite{AdaIN, InstanceNorm} to input latent information to the decoder.
Many techniques have been developed to reconsider AdaIN for different needs.
\cite{StyleGAN2} observed that using AdaIN inside the generator leads to blob artifacts and replaced it with a hypernetwork-based \cite{Hypernetworks} weights modulation.
\cite{SPADE} proposed to denormalize activations using embeddings extracted from a semantic mask.
\cite{SEAN} modified their approach by producing separate style codes for each semantic class and applied them on a per-region basis.
\cite{FUNIT} performed few-shot image-to-image translation from a given source domain via a mapping determined by the target's domain style space.
In our case, we do not use shifts and compute the scale weights by interpolating nearby latent codes using their coordinate positions instead of using global ones for the whole image as is done in the traditional AdaIN.

\textbf{Equivariant models}.
\cite{ConvsShiftInv} added averaging operation to a convolutional block to improve its invariance/equivariance to shifts.
\cite{NaturalEquivariance} explored natural equivariance properties inside the existing classifiers.
\cite{SphericalCNNs} developed a convolutional module equivariant to sphere rotations, and
\cite{GaugeEquivCNNs} generalized this idea to other symmetries.
In our case, we manually construct a model to be equivariant to shifts in the coordinate space.

\section{Method}\label{sec:method}

\begin{figure}
    \centering
    \includegraphics[width=\linewidth]{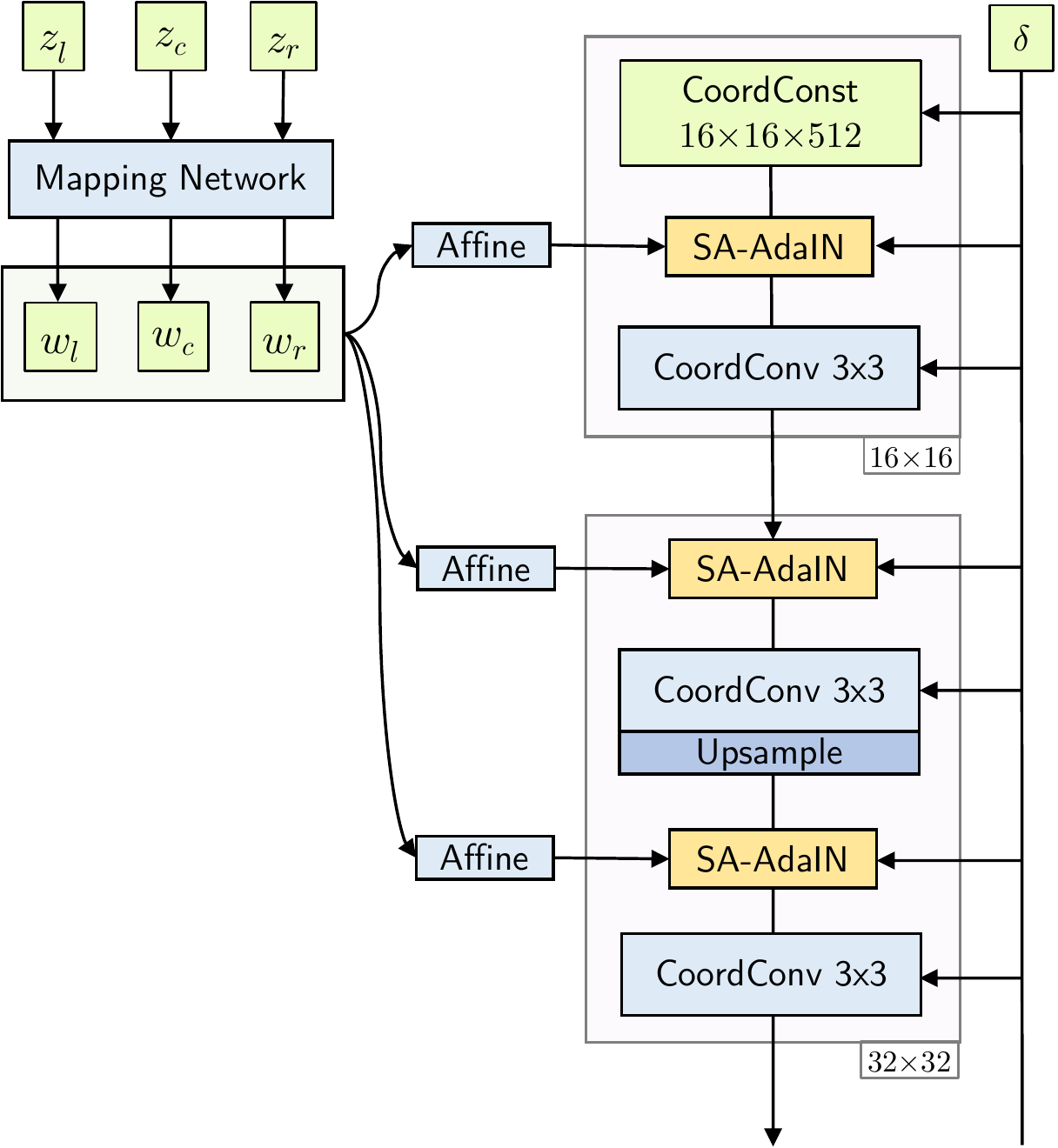}
    \caption{Illustration of our proposed generator. In this illustration, we omit some standard StyleGAN2 layers that are not essential for our architecture to not clutter the exposition. The full architecture is provided in Appendix \apref{ap:implementation-details}. \textsf{CoordConst} and \textsf{CoordConv3x3} are analogs of StyleGAN2's \textsf{Const} and \textsf{Conv3x3} blocks, but with coordinates embeddings concatenated to the hidden representations.}
    \label{fig:architecture}
\end{figure}

We build upon StyleGAN2 \cite{StyleGAN2} architecture and modify only its generator $\G$ as illustrated in Fig~\ref{fig:architecture}.
All the other components, including discriminator $\D$, the loss terms, and the optimization procedure, are left untouched.

To produce an image with our generator, we first need to sample \textit{anchors}: latent codes that define the context in the space region in which the image is located.
In this work, we consider only horizontal infinite image generation.
Thus we need only three anchors to define the context: left anchor $\bm w_l$, center anchor $\bm w_c$ and right anchor $\bm w_r$.
Following StyleGAN2, we produce a latent code $\bm w$ with the mapping network $\bm w \sim \F(\bm z)$ where $\bm z \sim \mathcal{N}(\bm 0, \bm I_{d_z})$.

\begin{figure}
    \centering
    \includegraphics[width=0.9\linewidth]{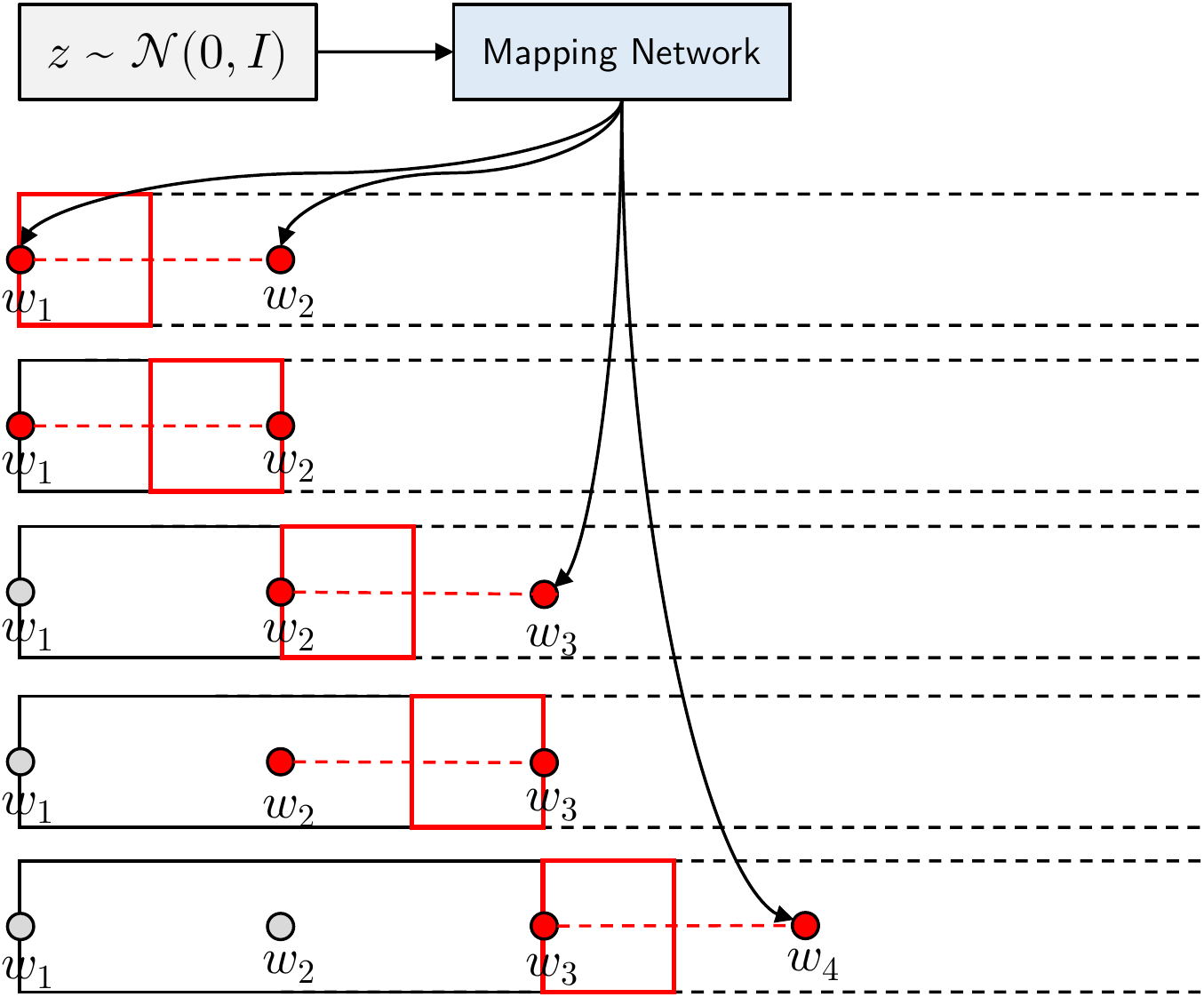}
    \caption{Inference process of our model at test time for $d = 2W$ (i.e. the distance between anchors is twice larger than the frame width). We sample new anchors $\bm{w}_i$ on the fly at positions $0, d, 2d, 3d, ...$. Since only relative positional information is provided to the decoder, we can decode images at any location $x \in (-\infty, \infty)$.}
    \label{fig:inference}
\end{figure}

To generate an image, we first need to define its location in space.
It is defined relative to the left anchor $\bm w_l$ by a position of its left border $\delta \in [0, 2\cdot d - W]$, where $d$ is the distance between anchors and $W$ is the frame width.
In this way, $\delta$ gives flexibility to position the image anywhere between $\bm w_l$ and $\bm w_r$ anchors in such a way that it lies entirely inside the region controlled by the anchors.
During training, $\bm w_l, \bm w_c, \bm w_r$ are positioned in the locations $-d, 0, d$ respectively along the $x$-axis and $\delta$ is sampled randomly.
At test time, we position anchors $\bm w_i$ at positions $0, d, 2d, 3d, ...$ and move with the step size of $\delta = W$ along the $x$-axis while generating new images.
This is illustrated in Figure~\ref{fig:inference}.

Traditional StyleGAN2 inputs a latent code into the decoder via an AdaIN-like \cite{AdaIN} weight demodulation mechanism, which is not suited for our setup since we use different latent codes depending on a feature coordinate position.
This forces us to modify it into Spatially-Aligned AdaIN (SA-AdaIN), which we describe in Sec~\ref{sec:method:sa-adain}.

\begin{figure}
    \centering
    \includegraphics[width=\linewidth]{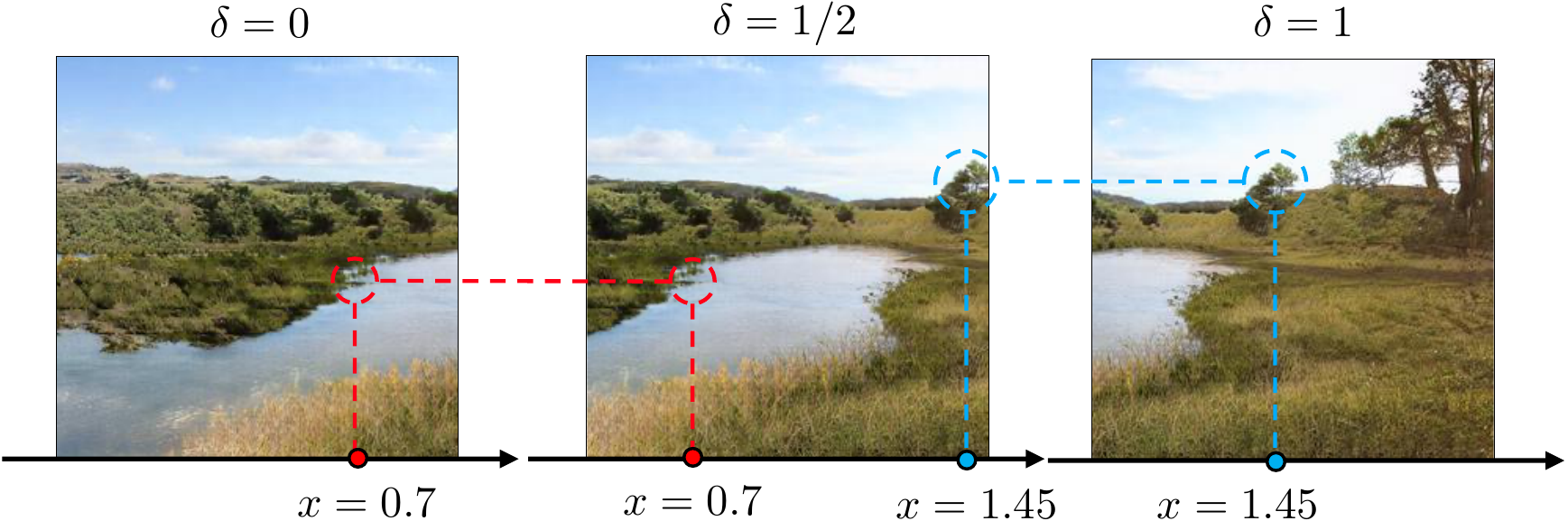}
    \caption{Our generator has the equivariance property by construction: we depict three samples with the coordinate shifts of 0, $1/2$ and 1, respectively, and this makes the resulted output move accordingly. As highlighted by dash circles, pixel values are equal when their coordinates are equal (up to numerical precision) for different samples. Samples remain of the same quality and diversity for any shift $s \in (-\infty, \infty)$.}
    \label{fig:equivariance}
\end{figure}

Our generator architecture is coordinate-based and, inspired by \cite{CocoGAN}, we generate an image as $16$ independent vertical patches, which are then concatenated together (see \apref{ap:implementation-details} for the illustration).
Independent generation is needed to make the generator learn how to stitch nearby patches using the coordinates information: at test-time, it will be stitching together an infinite amount of them.
Such a design also gives rise to an attractive property: spatial equivariance, that we illustrate in Figure~\ref{fig:equivariance}.
It arises from the fact that each patch does not depend on nearby patches, but only on the anchors $\bm w_l, \bm w_c, \bm w_r$ and its relative coordinates position $\delta$.

\subsection{Aligning Latent and Image Spaces}
\label{sec:alis}

The core idea of our model is positioning \textit{global} latent codes (\textit{anchors}) on the image coordinates grid and slowly varying them with linear interpolation while moving along the plane.
In this way, interpolation between the latent codes can be seen as the camera translation between two unrelated scenes corresponding to different independent anchors.
We call those anchors \textit{global} because each one of them influences many frames.
The idea for 1D alignment (i.e., where we move only along a single axis) is illustrated on Figure~\ref{fig:alignment}.

Imagine that we need to generate a pixel or intermediate feature value $\bm v$ at position $(x, y) \in \R^2$.
A traditional generator would produce a latent code $\bm w$ and generate the value based on it: $\bm v(x, y) = \G(x, y; \bm w)$.
But in our case, we make the latent code be dependent on $x$ (it's trivial to generalize it to be dependent on both $x$ and $y$):
\begin{equation}
\bm v(x, y) = \G(x, y; \bm w_x),
\end{equation}
where $\bm w_x$ is computed as a linear interpolation between the nearby anchors $\bm w_a$ and $\bm w_b$ at positions $a, b \in \R$:
\begin{equation}
\bm w_x = \alpha \bm w_a + (1 - \alpha) \bm w_b,
\end{equation}
and $\alpha$ is the normalized distance from $x$ to $b$: $\alpha = (b-x)/(b-a)$.

In this way, latent and image spaces become aligned with one another: any movement in the coordinate space spurs movement in the latent space and vice versa.
By training $\G$ to produce images in the interpolated regions, it implicitly learns to connect $\bm w_a$ and $\bm w_b$ with realistic inter-scene frames.
At test time, this allows us to connect any two independent $\bm w_a$ and $\bm w_b$ into a single panorama by generating in-between frames, as can be seen in Figure~\ref{fig:teaser}.

\subsection{Spatially Aligned AdaIN (SA-AdaIN)}
\label{sec:method:sa-adain}

Our $\G$ architecture is based on StyleGAN2's generator, which uses an AdaIN-like mechanism of modulating the decoder's convolutional weights. This mechanism is not suited for our setup since one cannot make convolutional weights be dependent on the coordinate position efficiently.
That is why we develop a specialized AdaIN variation that can be implemented efficiently on modern hardware.

Classical AdaIN \cite{AdaIN} works the following way: given an input $\bm{h} \in \R^{c \times s^2}$ of resolution $s^2$ (for simplicity, we consider it to be square), it first normalizes it across spatial dimensions and then rescales and shifts with parameters $\bm{\gamma}, \bm\beta \in \R^c$:
\begin{equation}
\text{AdaIN}(\bm h, \bm \gamma, \bm \beta) = \bm\gamma \cdot \frac{\bm h - \bm\mu(\bm h)}{\bm\sigma(\bm h)} + \bm\beta
\end{equation}
where $\bm\mu(\bm h), \bm\sigma(\bm h) \in \R^c$ are mean and standard deviation computed across the spatial axes and all the operations are applied element-wise.
It was recently shown that the performance does not degrade when one removes the shifting \cite{StyleGAN2}:
\begin{equation}
\text{AdaIN}'(\bm h, \bm \gamma) = \bm\gamma \cdot \bm h / \bm\sigma(\bm h)
\end{equation}
thus we build on top of this simplified version of AdaIN.

Our Spatially-Aligned AdaIN (SA-AdaIN) is an analog of AdaIN' for a scenario where latent and image spaces are aligned with one other (as described in Sec~\ref{sec:alis}), i.e. where the latent code is different depending on the coordinate position we compute it in.
This section describes it for 1D-case, i.e., when the latent code changes only across the horizontal axis, but our exposition can be easily generalized to the 2D case (actually, for any $N$-D).

While generating an image, ALIS framework uses $\bm w_l, \bm w_c$ and $\bm w_r$ to compute the interpolations $\bm w_x$ for the required positions $x$.
However, following StyleGAN2, we input to the convolutional layers not the ``raw'' latent codes $\bm w$, but style vectors $\bm\gamma$ obtained through an affine transform $\bm\gamma = \bm{A}^\ell\bm w + \bm b^\ell$ where $\ell$ denotes the layer index.
Since a linear interpolation and an affine transform are interchangeable, from the performance considerations we first compute $\bm\gamma_l, \bm\gamma_c, \bm\gamma_r$ from $\bm w_l, \bm w_c, \bm w_r$ and then compute the interpolated style vectors $\bm\gamma_x$ instead of interpolating $\bm w_l, \bm w_c, \bm w_r$ into $\bm w_x$ immediately.

SA-AdaIN works the following way.
Given anchor style vectors $\bm\gamma_l, \bm\gamma_c, \bm\gamma_r$, image offset $\delta$, distance $d$ between the anchors and the resolution of the hidden representation $s$, it first computes the grid of interpolated styles $\bm\Gamma = [\bm\gamma_1, \bm\gamma_2, ..., \bm\gamma_s] \in \R^{s \times c}$, where:
\begin{equation}
\bm\gamma_k = 
\begin{cases}
\frac{d - \delta - k/s}{d}\bm\gamma_l + \frac{\delta + k/s}{d}\bm\gamma_c, & \text{if }\delta + k/s > d \\
\frac{2d - \delta - k/s}{d}\bm\gamma_c + \frac{\delta + k/s - d}{d}\bm\gamma_r, & \text{otherwise}
\end{cases}
\end{equation}
This formulation assumes that anchors $\bm\gamma_l, \bm\gamma_c, \bm\gamma_r$ are located at positions $-d, 0, d$ respectively.

Then, just like AdaIN', it normalizes $\bm h$ to obtain $\tilde{\bm h} = \bm h / \bm\sigma(\bm h)$.
After that, it element-wise multiplies $\bm\Gamma$ and $\bm h$, broadcasting the values along the vertical positions:
\begin{equation}
[\text{SA-AdaIN}(\bm x, \bm\gamma_l, \bm\gamma_c, \bm\gamma_r, \delta)]_k = \bm\gamma_k \cdot [\bm h / \bm\sigma(\bm h)]_k,
\end{equation}
where we denote by $[\cdot]_k$ the $k$-th vertical patch of a variable of size $s \times 1 \times c$.
Note that since our $\G$ produces images in a patch-wise fashion, we normalize across patches instead of the full images.
Otherwise, it will break the equivariance and lead to seams between nearby frames.

\begin{figure}
    \centering
    \begin{subfigure}[b]{\linewidth}
        \centering
        \includegraphics[width=\textwidth]{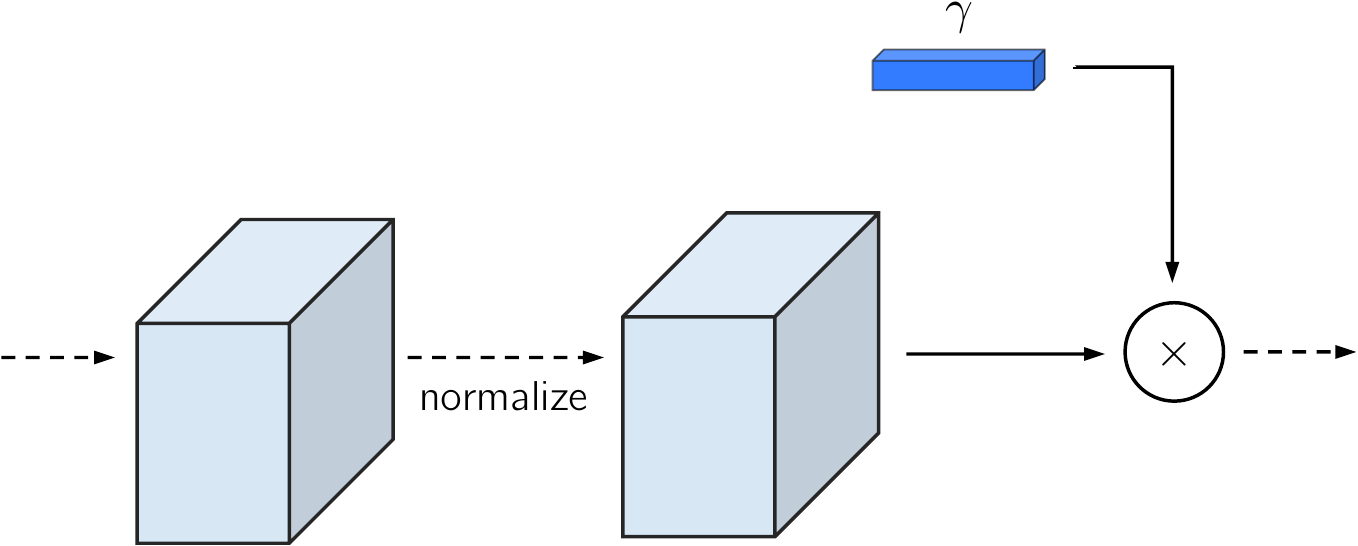}
        \caption{AdaIN without shifting \cite{StyleGAN2}}
    \end{subfigure}
    \begin{subfigure}[b]{\linewidth}
        \centering
        \vspace{0.5cm}
        \includegraphics[width=\textwidth]{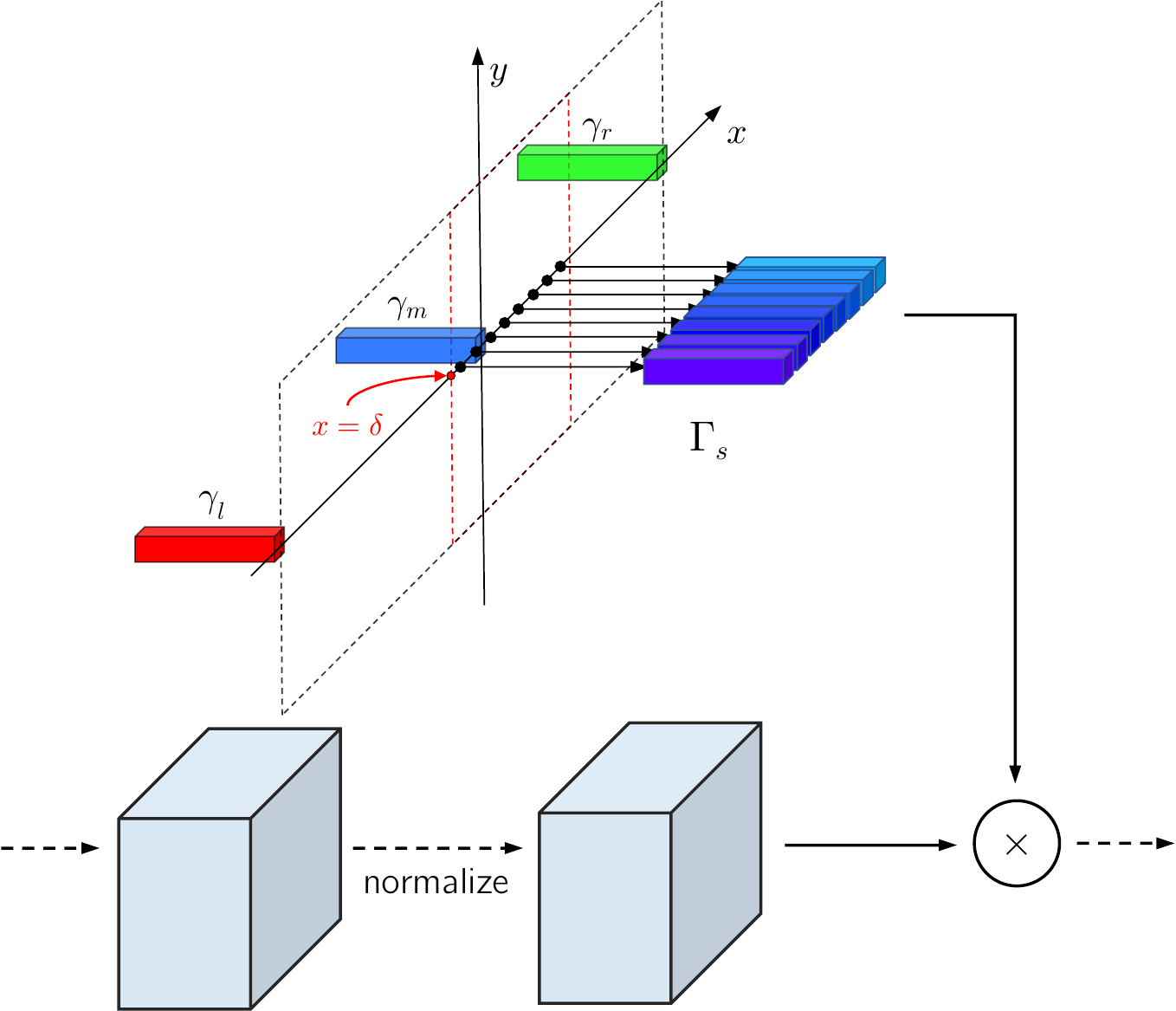}
        \caption{Spatially-Aligned AdaIN}
    \end{subfigure}
    \caption{Top: AdaIN \cite{AdaIN} without shifting (as explored in \cite{StyleGAN2}). Bottom: Spatially-Aligned AdaIN (SA-AdaIN). Style vectors $\bm\gamma_\ell, \bm\gamma_c, \bm\gamma_r$ are positioned on the 2D coordinate space and we compute a style vector in each location as a linear interpolation between its neighbors. For both AdaIN and SA-AdaIN, we broadcast styles' dimensions in the multiplication operation to match the input tensor shape.}
    \label{fig:sa-adain}
\end{figure}

SA-AdaIN is illustrated in Figure~\ref{fig:sa-adain}.

\begin{figure*}
    \centering
    \begin{subfigure}[b]{\linewidth}
        \centering
        \includegraphics[width=\textwidth]{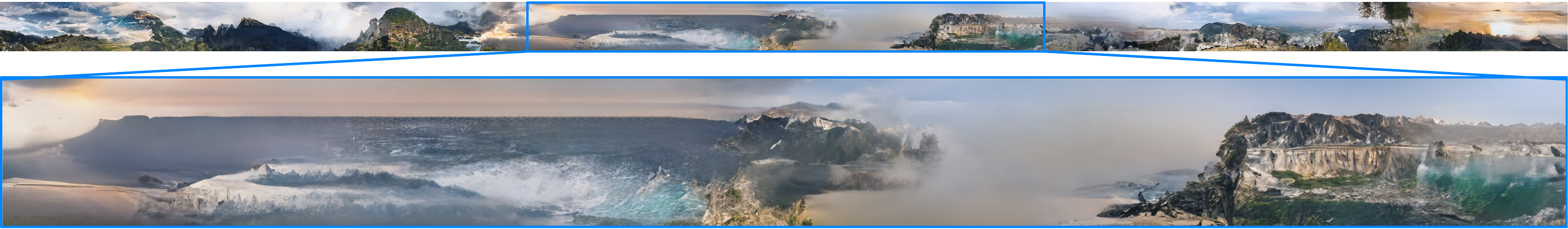}
        \includegraphics[width=\textwidth]{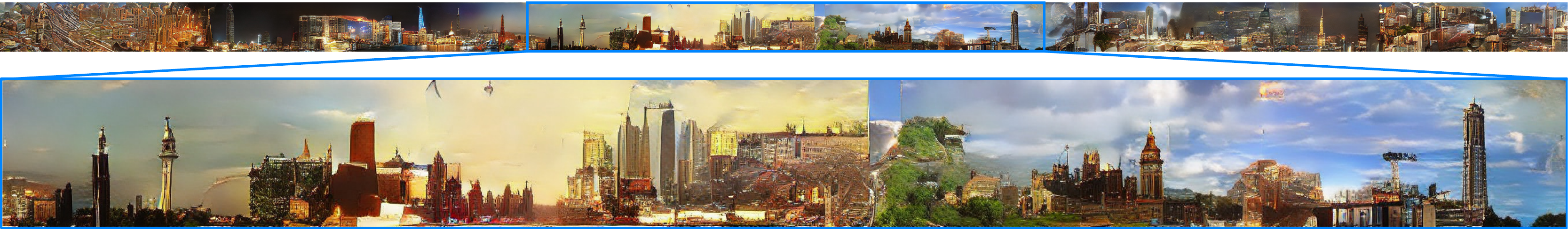}
        \caption{Taming Transformers \cite{TamingTransformers} for \textit{unconditional} generation (the original paper mainly focused on the conditional generation from semantic masks/depth maps/etc).}
    \end{subfigure}
    \begin{subfigure}[b]{\linewidth}
        \centering
        \includegraphics[width=\textwidth]{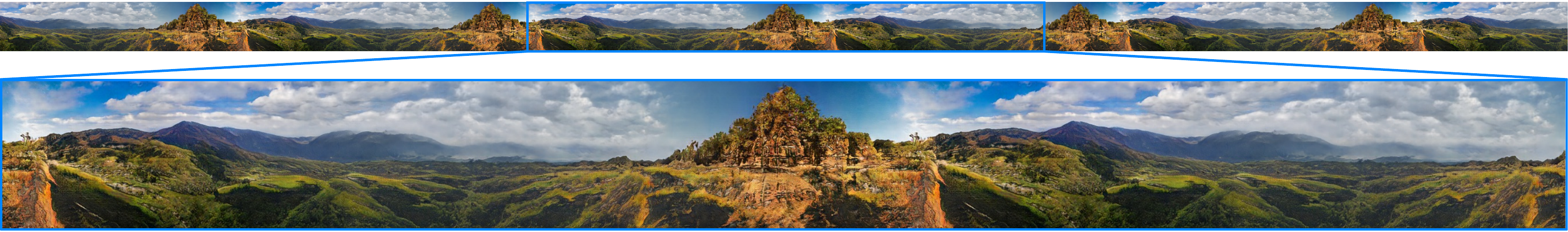}
        \includegraphics[width=\textwidth]{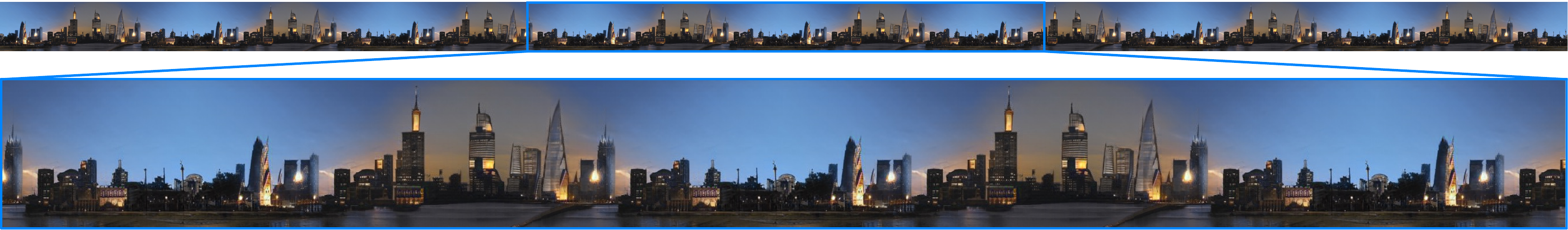}
        \caption{LocoGAN \cite{LocoGAN} in the StyleGAN2 \cite{StyleGAN2} framework + Fourier positional embeddings \cite{SIREN, FourierINR}}
    \end{subfigure}
    \begin{subfigure}[b]{\linewidth}
        \vspace{0.5cm}
        \centering
        \includegraphics[width=\textwidth]{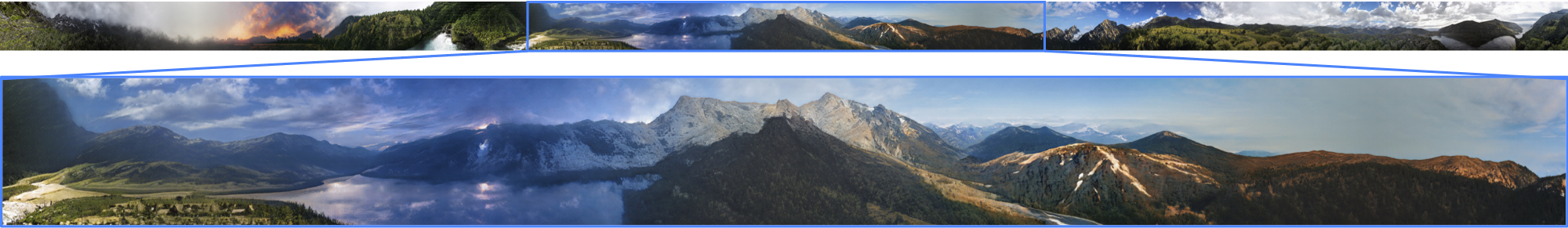}
        \includegraphics[width=\textwidth]{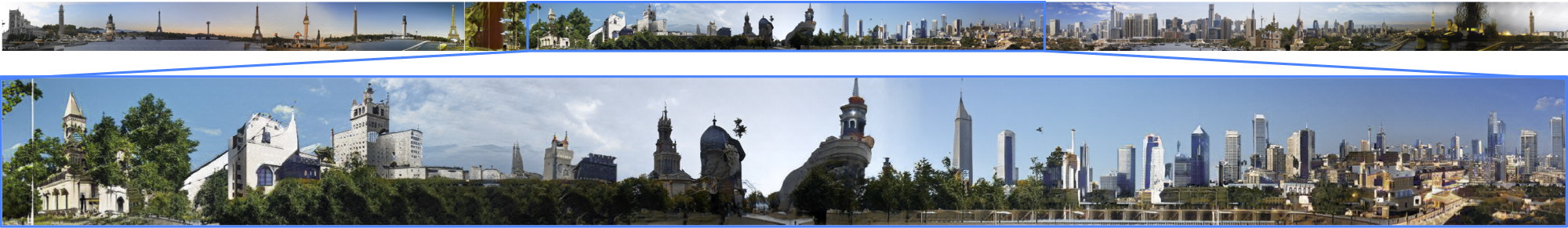}
        \caption{ALIS \ours}
    \end{subfigure}
    \caption{Qualitative comparison between different methods on LHQ and LSUN Tower. More samples are in Appendix~\apref{ap:samples}.}
    \label{fig:qualitative-comparison}
\end{figure*}

\subsection{Which datasets are ``connectable''?}
\label{sec:method:connectable-data}

As mentioned in \cite{TamingTransformers}, to generate arbitrarily-sized images, we want the data statistics to be invariant to their spatial location in an image.
This means that given an image patch, one should be unable to confidently predict which part of an image it comes from.
However, many datasets either do not have this property (like FFHQ \cite{StyleGAN}) or have it only for a small number of images.
To check if images in a given dataset have spatially invariant statistics and to extract a subset of such images, we developed the following simple procedure.
Given a dataset, we train a classifier on its patches to predict what part of an image a patch is coming from.
If a classifier cannot do this easily (i.e., it has low accuracy for such predictions), then the dataset does have spatially invariant statistics.
To extract a subset of such good images, we measure this classifier's confidence on each image and select those for which its confidence is low.
This procedure is discussed in details in Appendix~\apref{ap:spatial-inv}.

\section{Landscapes HQ dataset}\label{sec:lhq}
We introduce Landscapes HQ (LHQ): a dataset of 90k high-resolution ($\geq 1024^2$) nature landscapes that we crawled and preprocessed from two sources: Unsplash (60k) and Flickr (30k).
We downloaded 500k of images in total using a list of manually constructed 450 search queries and then filtered it out using a blacklist of 320 image tags.
After that, we ran a pretrained Mask R-CNN to remove the pictures that likely contain objects on them.
As a result, we obtained a dataset of 91,693 high-resolution images.
The details are described in Appendix~\apref{ap:lhq}.

\section{Experiments}\label{sec:experiments}



\begin{table*}
\caption{Scores for different models on different datasets in terms of FID and $\infty$-FID. ``N/A'' denotes ``not-applicable''.}
\label{table:main-results}
\centering
\begin{tabular}{|l|ll|ll|ll|l|c|}
\hline
\multirow{2}{*}{Method} & \multicolumn{2}{c|}{Bridge $256^2$} & \multicolumn{2}{c|}{Tower $256^2$} & \multicolumn{2}{c|}{Landscapes $256^2$} & \multirow{2}{*}{\hspace{1.4em}Speed} & \multirow{2}{*}{\#params} \\
& FID & $\infty$-FID & FID & $\infty$-FID & FID & $\infty$-FID &  & \\
\hline
Taming Transformers~\cite{TamingTransformers} & 56.06 & 58.27 & 50.16 & 51.32 & 61.95 & 64.3 & 9981 ms/img & 377M\\
LocoGAN~\cite{LocoGAN}+SG2+Fourier & 9.02 & 264.7 & 8.36 & 381.1 & 7.82 & 211.2 & 74.7 ms/img & 53.7M \\
ALIS \ours & 10.24 & 10.79 & 8.83 & 8.99 & 10.48 & 10.64 & 53.9 ms/img & 48.3M \\
~~~w/o coordinates & 13.21 & 13.92 & 10.32 & 10.17 & 12.63 & 13.07 & 46.8 ms/img & 47.1M \\
\hline
\hline
StyleGAN2 (\textsf{config-e}) & 7.33 & N/A & 6.75 & N/A & 3.94 & N/A & 32.4~ms/img & 47.1M \\
\hline
\end{tabular}
\end{table*}

\begin{figure}
    \centering
    \includegraphics[width=0.8\linewidth]{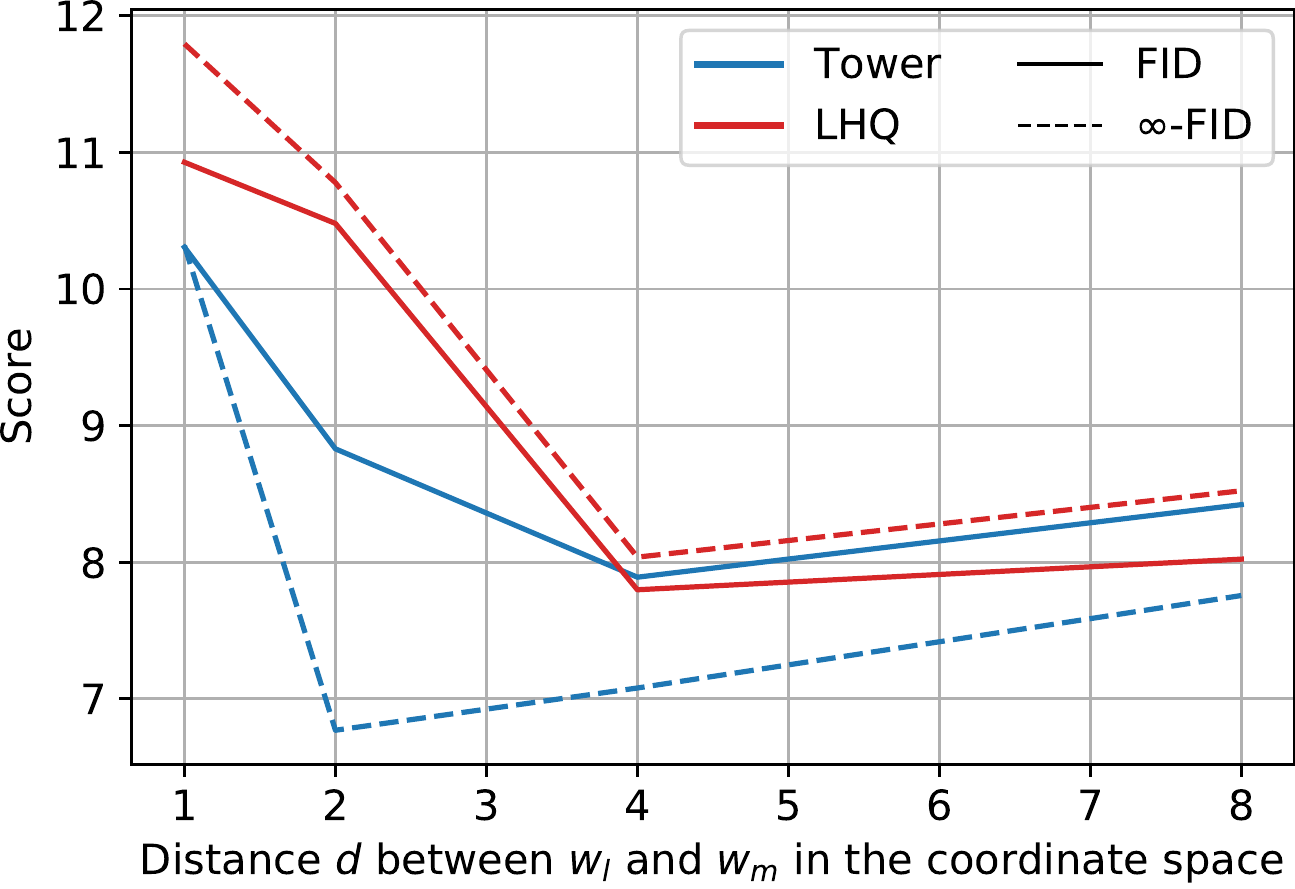}
    \caption{Varying distance $d$ between the anchors for LSUN Tower and LHQ datasets. Larger distance leads to better per-frame image quality, but produces repetitions artifacts as depicted on Figure~\ref{fig:large-dist-problems}.}
    \label{fig:w_coord_dists}
\end{figure}

\begin{figure}
    \centering
    \includegraphics[width=\linewidth]{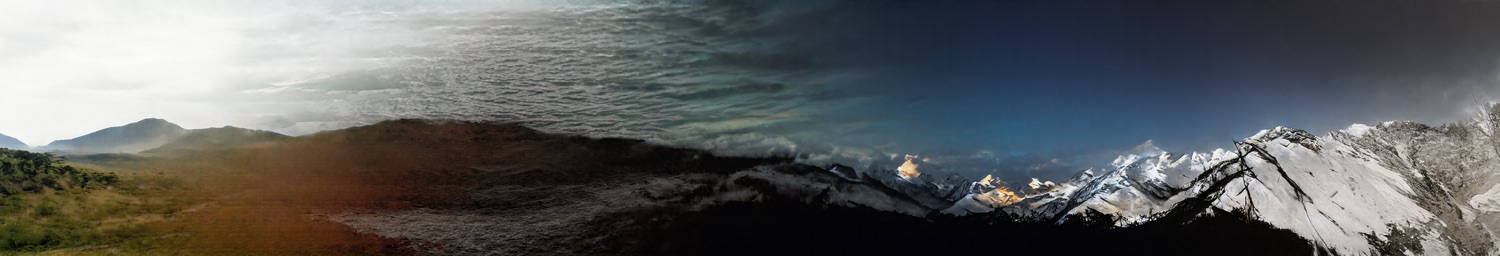}
    \includegraphics[width=\linewidth]{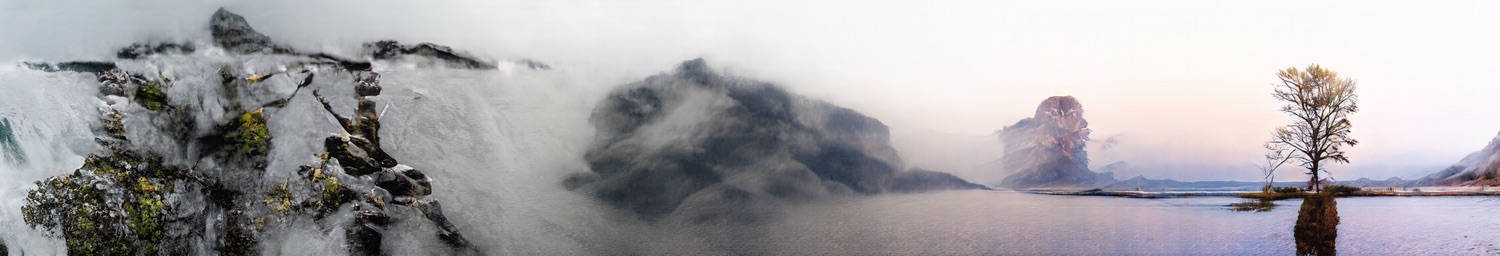}
    \includegraphics[width=\linewidth]{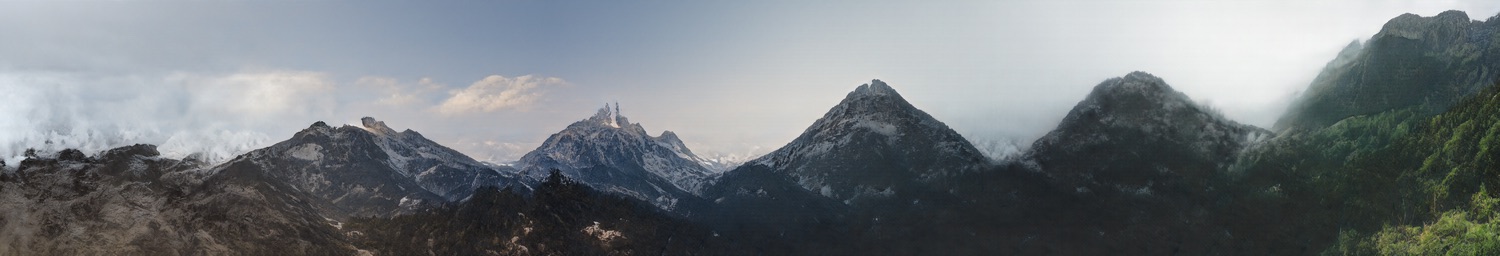}
    \includegraphics[width=\linewidth]{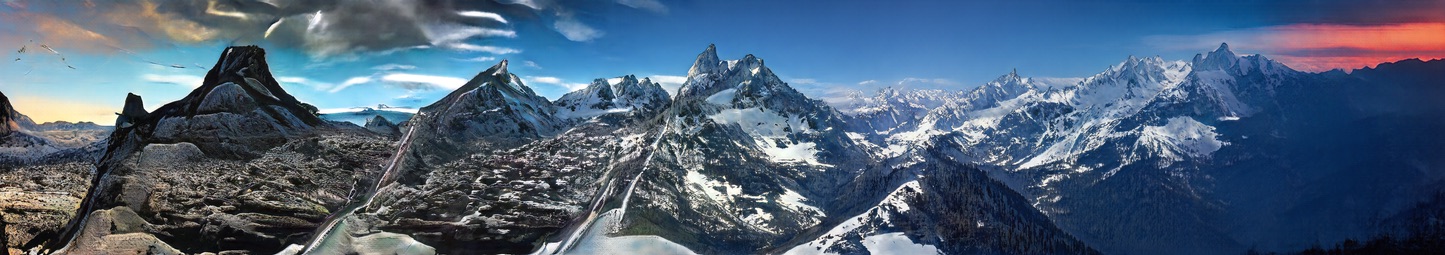}
    \caption{Failure cases of our method: 1) top 2 rows --- connecting too different anchors (like close-by water and far-away mountains); and 2) bottom 2 rows --- content repetitions (which arise due to the usage of periodic positional embeddings \cite{FourierINR, SIREN}).}
    \label{fig:failure-cases}
\end{figure}

\begin{figure}
    \centering
    \includegraphics[width=\linewidth]{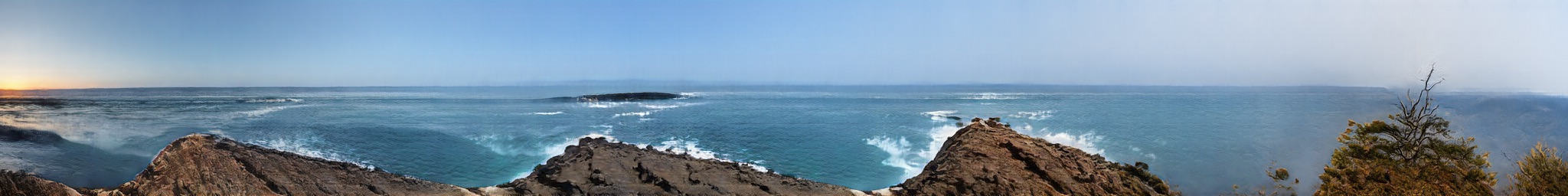}
    \includegraphics[width=\linewidth]{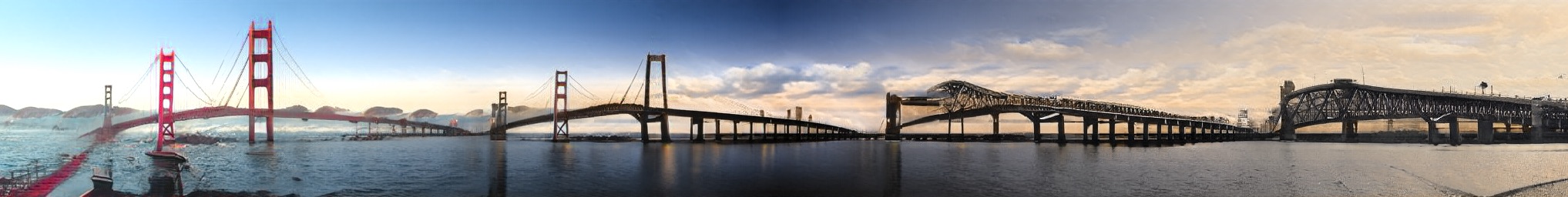}
    \caption{A problem of using large distance $d$ between the anchors (in the above case, model was trained with $d = 16$). Though it improves per-frame image quality, the model starts repeating itself during the generation process.}
    \label{fig:large-dist-problems}
\end{figure}

\begin{figure}
    \centering
    \includegraphics[width=\linewidth]{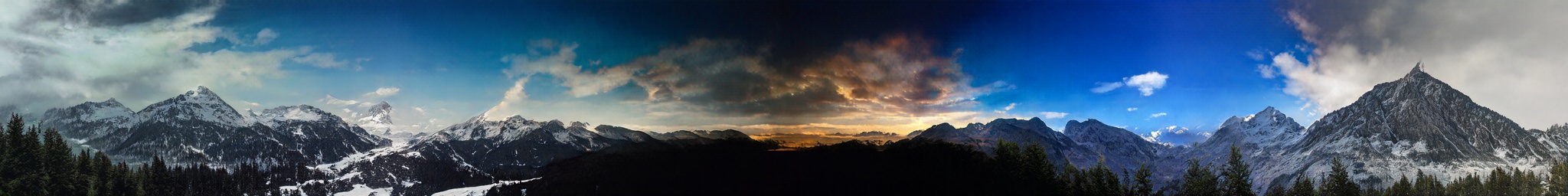}
    \includegraphics[width=\linewidth]{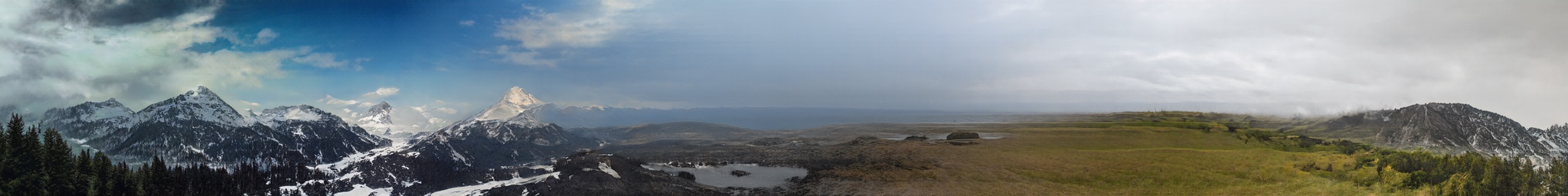}
    \includegraphics[width=\linewidth]{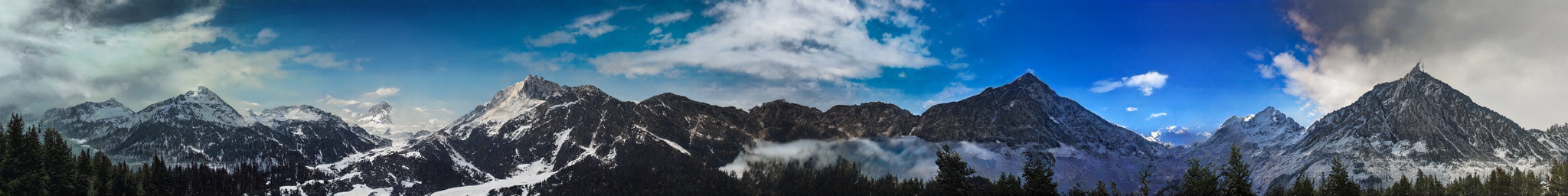}
    \includegraphics[width=\linewidth]{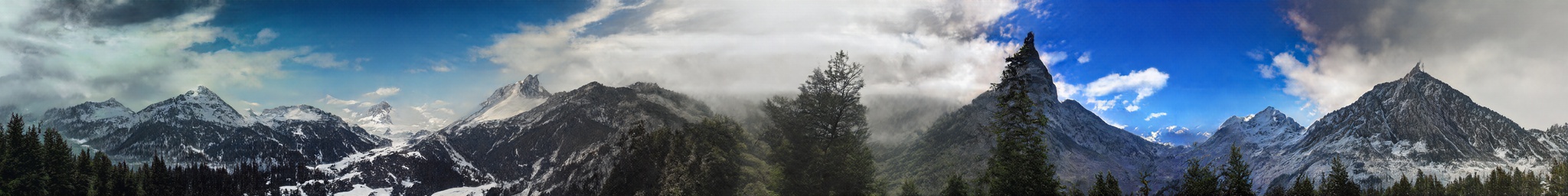}
    \includegraphics[width=\linewidth]{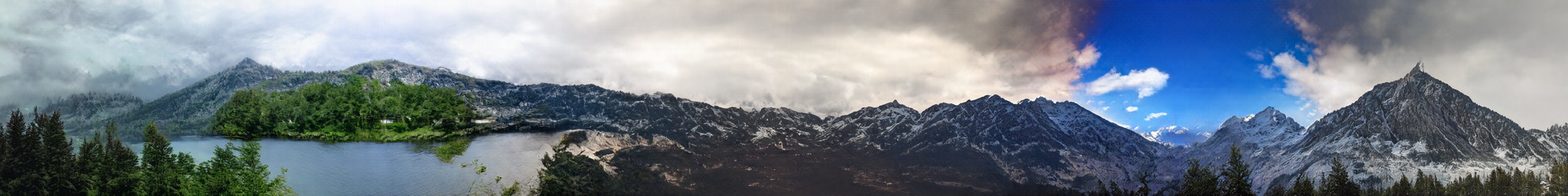}
    \caption{ALIS allows to resample any part of an image without breaking its ``connectivity'': frames are still consistent both locally and globally.}
    \label{fig:resampling}
\end{figure}

\textbf{Datasets}.
We test our model on 4 datasets: LSUN Tower $256^2$, LSUN Bridge $256^2$ and LHQ $256^2$ (see Sec~\ref{sec:lhq}).
We preprocessed each dataset with the procedure described in Algorithm~\apref{alg:spatial-inv} in Appendix~\apref{ap:spatial-inv} to extract a subset of data with (approximately) spatially invariant statistics.
We emphasize that we focus on infinite horizontal generation, but the framework is easily generalizable for the joint vertical+horizontal infinite generation.
We also train ALIS on LHQ $1024^2$ to test how it scales to high resolutions.

\textbf{Evaluation}.
We use two metrics to compare the methods: a popular FID measure \cite{TTUR} and our additionally proposed $\infty$-FID, which is used to measure the quality and diversity of an ``infinite'' image.
It is computed in the following way.
For a model with the output frame size of $256^2$, we first generate a very wide image of size $256 \times (50000 \cdot 256)$.
Then, we slice it into $50000$ frames of size $256^2$ without overlaps or gaps and compute FID between the real dataset and those sliced frames.
We also compute a total number of parameters and inference speed of the generator module for each baseline.
Inference speed is computed as a speed to generate a \textit{single} frame of size $256^2$ on NVidia V100 GPU.

\textbf{Baselines}.
For our main baselines, we use two methods: Taming Transformers (TT) \cite{TamingTransformers} and LocoGAN \cite{LocoGAN}.
For LocoGAN, since by default it is a small model (5.5M parameters vs {$\approx$}50M parameters in StyleGAN2) and does not employ any StyleGAN2 tricks that boost the performance (like style mixing, equalized learning rate, skip/resnet connections, noise injection, normalizations, etc.) we reimplemented it on top of the StyleGAN2 code to make the comparison fair.
We also replaced raw coordinates conditioning with Fourier positional embeddings since raw coordinates do not work for $(-\infty, \infty)$ range and were recently shown to be inferior \cite{SIREN, FourierINR, INR_GAN, CIPS}.
We called this model LocoGAN+SG2+Fourier.
Besides the methods for infinite image generation, we compute the performance of the traditional StyleGAN2 as a lower bound on the possible image generation quality on a given dataset.

Each model was trained on 4 v100 GPUs for 2.5 days, except for TT, which was trained for 5 days since it is a two-stage model: 2.5 days for VQGAN and then 2.5 days for Transformer.
For TT, we used the official implementation with the default hyperparameters setup for unconditional generation training.
See Appendix~\apref{ap:taming-transformers} for the detailed remarks on the comparison to TT.
For StyleGAN2-based models, we used \textsf{config-e} setup (i.e. half-sized high-resolution layers) from the original paper \cite{StyleGAN2}.
We used precisely the same training settings (loss terms, optimizer parameters, etc.) as the original StyleGAN2 model.

\textbf{Ablations}.
We also ablate the model in terms of how vital the coordinates information is and how distance between anchors influences generation.
For the first part, we replace all \textsf{Coord-*} modules with their non-coord-conditioned counterparts.
For the second kind of ablations, we vary the distance between the anchors in the coordinate space for values $d = 1, 2, 4, 8$.
This distance can also be understood as an aspect ratio of a single scene.

\textbf{Results}.
The main results are presented in Table~\ref{table:main-results}, and we provide the qualitative comparison on Figure~\ref{fig:qualitative-comparison}.
To measure $\infty$-FID for TT, we had to simplify the procedure since it relies on GroupNorm in its decoder and generated 500 images of width $256 \times 100$ instead of a single image of width $256 \times 50000$.
We emphasize, that it is an \textit{easier} setup and elaborate on this in Appendix~\ref{ap:taming-transformers}.
Note also, that the TT paper \cite{TamingTransformers} mainly focused on \textit{conditional} image generation from semantic masks/depths maps/class information/etc, that's why the visual quality of TT's samples is lower for our \textit{unconditional} setup.

The $\infty$-FID scores on Table~\ref{table:main-results} demonstrate that our proposed approach achieves state-of-the-art performance in the generation of infinite images.
For the traditional FID measured on independent frames, its performance is on par with LocoGAN.
However, the latter completely diverges in terms of infinite image generation because spatial noise does not provide global variability, which is needed to generate diverse scenes.
Moreover, we noticed that LocoGAN has learned to ignore spatial noise injection \cite{StyleGAN} to make it easier for the decoder to stitch nearby patches, which shuts off its only source of scene variability.
For the model without coordinates conditioning, image quality drops: visually, they become blurry (see Appendix~\apref{ap:samples}) since it is harder for the model to distinguish small shifts in the coordinate space.
When increasing the coordinate distance between anchors (the width of an image equals 1 coordinate unit), traditional FID improves.
However, this leads to repetitious generation, as illustrated in Figure~\ref{fig:large-dist-problems}.
It happens due to short periods of the periodic coordinate embeddings and anchors changing too slowly in the latent space.
When most of the positional embeddings complete their cycle, the model has not received enough update from the anchors' movement and thus starts repeating its generation.
The model trained on $1024^2$ crops of LHQ achieves FID/$\infty$-FID scores of 10.11/10.53 respectively and we illustrate its samples in Figure~\ref{fig:teaser} and Appendix~\apref{ap:samples}.

As depicted in Figure~\ref{fig:failure-cases}, our model has two failure modes: sampling of \textit{too} unrelated anchors and repetitious generation.
The first issue could be alleviated at test-time by using different sampling schemes like truncation trick \cite{BigGAN, StyleGAN2} or clustering the latent codes.
A more principled approach would be to combine autoregressive inference of TT \cite{TamingTransformers} with our ideas, which we leave for future work.

One of our model's exciting properties is the ability to replace or swap any two parts of an image without breaking neither its local nor global consistency.
It is illustrated in Figure~\ref{fig:resampling} where we resample different regions of the same landscape.
For it, the middle parts are changing, while the corner ones remain the same.

\section{Conclusion}

In this work, we proposed an idea of aligning the latent and image spaces and employed it to build a state-of-the-art model for infinite image generation.
We additionally proposed a helpful $\infty$-FID metric and a simple procedure to extract from any dataset a subset of images with approximately spatially invariant statistics.
Finally, we introduced LHQ: a novel computer vision dataset consisting of 90k high-resolution nature landscapes.

{\small
\bibliographystyle{ieee_fullname}
\bibliography{egbib}
}

\clearpage
\newpage

\appendix
\onecolumn
\section{Discussion}

\subsection{Limitations}

Our method has the following limitations:
\begin{enumerate}
    \item Shift-equivariance of our generator is limited only to the step size equal to the size of a patch, i.e., one cannot take a step smaller than the patch size. In our case, it equals $16$ pixels for $256^2$-resolution generator and 64 pixels for $1024^2$-resolution one. In this way, our generator is \textit{periodic} shift-equivariant \cite{ConvsShiftInv}.
    \item As shown in \ref{fig:failure-cases}, connecting too different scenes leads to seaming artifacts. There are two ways to alleviate this: sampling from the same distribution mode (after clustering the latent space, like in Figure~\ref{fig:clustered-sampling}) and increasing the distance between the anchors. The latter, however, leads to repetitions artifacts, like in \ref{fig:large-dist-problems}.
    \item As highlighted in \ref{sec:method:connectable-data}, infinite image generation makes sense only for datasets of images with spatially invariant statistics and our method exploits this assumption. In this way, if one does not preprocess LSUN Bedroom with Algorithm~\ref{alg:spatial-inv}, then the performance of our method will degrade by a larger margin compared to TT \cite{TamingTransformers} for \textit{non-infinite} generation. However, after preprocessing the dataset, our method can learn to connect those ``unconnectable'' scenes (see Figure~\ref{fig:connecting-lsun-bedroom}). This limitation \textit{will not} be an issue in practice because when one is interested in non-infinite image generation, then a non-infinite image generator is employed. And when on tries to learn infinite generation on a completely ``unconnectable'' dataset, then any method will fail to do so (unless some external knowledge is employed to rule out the discrepancies).
    
\begin{figure}[h]
    \centering
    \includegraphics[width=\linewidth]{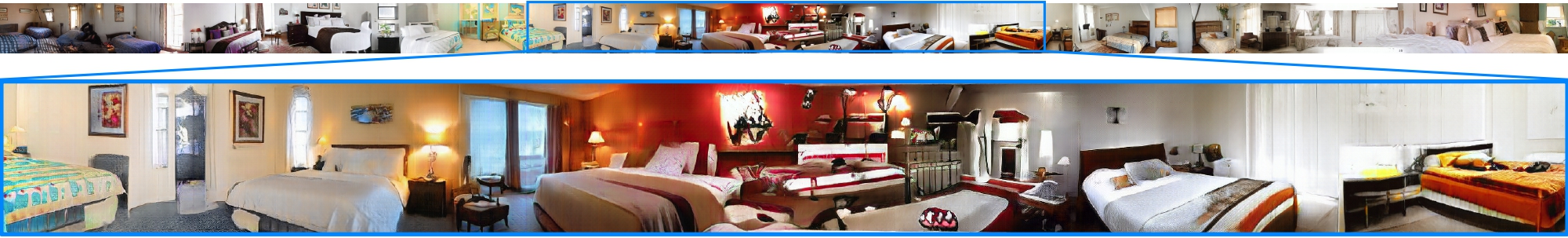}
    \\
    \vspace{1em}
    \includegraphics[width=\linewidth]{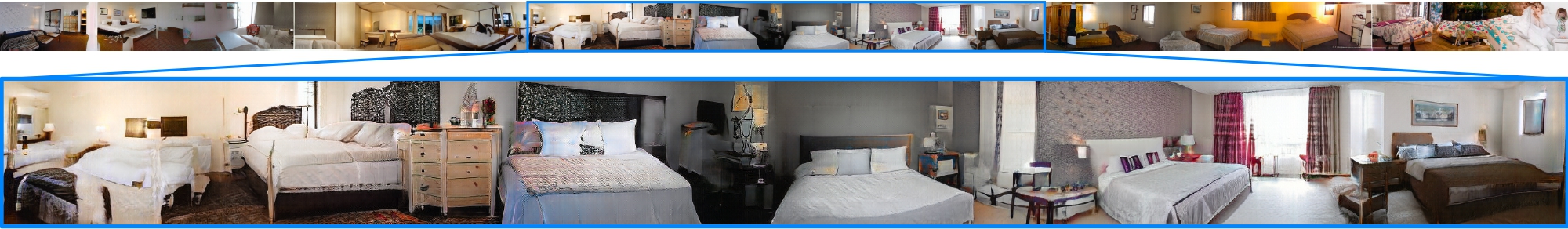}
    \caption{Connecting the ``unconnectable'' scenes of LSUN Bedroom. Though LSUN Bedroom is a dataset of images with spatially non-invariant statistics (see Appendix~\apref{ap:spatial-inv}), our method is still able to reasonably connect its scenes after the dataset is preprocessed with Algorithm~\ref{alg:spatial-inv}.}
    \label{fig:connecting-lsun-bedroom}
\end{figure}

    \item Preprocessing the dataset with our proposed procedure in Section~\ref{sec:method:connectable-data} changes the underlying data distribution since it filters away images with spatially non-invariant statistics. Note that we used the same processed datasets for our baselines in all the experiments to make the comparison fair.
\end{enumerate}

\subsection{Questions and Answers}

\newcommand{\Q}[1]{\vspace{0.5em}\textbf{Q:}~\textit{#1}}
\newcommand{\A}{\textbf{A:~}}

\vspace{-0.5em}

\vspace{-0.5em}\Q{What is spatial equivariance? Why is it useful? Is it unique to your approach, or other image generators also have it?}

\A We call a decoder \textit{spatially-equivariant} (shift-equivariant) if shifting its input results in an equal shift of the output.
It is not to be confused with spatial \textit{invariance}, where shifting an input does not change the output.
The equivariance property is interesting because it is a natural property that one seeks in a generator, i.e., by design, we want a decoder to move in accordance with the coordinate space movements.
It was also shown that shift-equivariance improves the performance \cite{ConvsShiftInv} of both decoder and encoder modules.
Traditional generators do not have it due to upsampling \cite{ConvsShiftInv}.
While we have upsampling procedures too, in our case they do not break \textit{periodic} shift-equivariance because they are patchwise.

\Q{Why did you choose the grid size of 16?}

\begin{figure}
    \centering
    \includegraphics[width=0.4\textwidth]{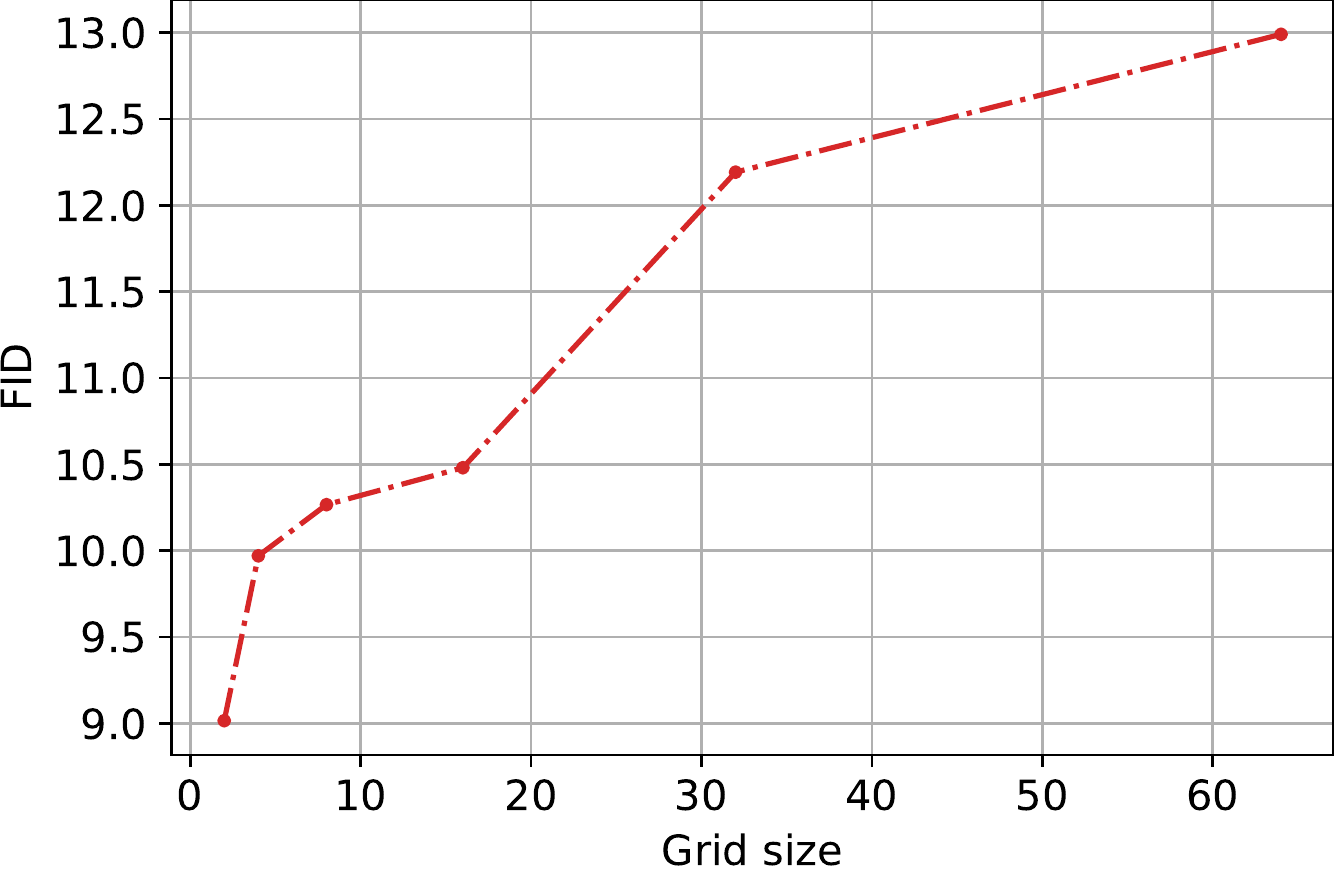}
    \caption{Ablating the grid size of the ALIS generator on LHQ $256^2$ for patchwise generation. Reducing the grid size improves the performance, but in the expense of not being able to do small steps in the coordinate space at test time. I.e. with $2 \times 2$ grid, one would achieve the highest FID score but will be able to generate an image only by large chunks (of size $W/2$) which might be undesirable.}
    \label{fig:grid-size-ablation}
\end{figure}

\A The grid size of 16 allows to perform small enough steps in the coordinate space while not losing the performance too much (as also confirmed by \cite{CocoGAN}).
Small steps in the coordinate space are needed from a purely design perspective to generate an image in smaller portions.
In the ideal case, one would like to generate an image pixel-by-pixel, but this extreme scenario of having $1\times1$ patches worsens image quality \cite{INR_GAN}.
We provide the ablation study for the grid size in Figure~\ref{fig:grid-size-ablation}.

\Q{Why didn't you compare against other methods for infinite image generation?}

\A There are two reasons for it. First, to the best of our knowledge, there are no infinite generators of \textit{semantically complex} images, like LSUN Tower or nature landscapes --- only texture-like images.
This is why we picked ``any-aspect-ratio'' image generators as our baselines: LocoGAN~\cite{LocoGAN} and Taming Transformers~\cite{TamingTransformers}.
Second, our LocoGAN+SG2+Fourier baseline encompasses the previous developments in infinite texture synthesis (like \cite{SpatialGAN, PS_GAN, InfinityGAN}).
It has the same architecture as the state-of-the-art PS-GAN \cite{PS_GAN}, but is tested not only on textures, but also semantically complex datasets, like LSUN Bedrooms and FFHQ.
See Table~\ref{table:benchmarks-comparison} for the benchmarks comparison in terms of features.

\begin{table}
\caption{Comparison of existing infinite image generators. We used LocoGAN~\cite{LocoGAN} as our benchmark since it shares all the features of previously developed infinite-texture image generators and was additionally tested on non-texture datasets. N/A denotes ``not applicable'': while $\infty$-GAN was trained on non-texture images, its generations are in the ``texture-like'' spirit.}
\label{table:benchmarks-comparison}
\centering
\resizebox{1.0\linewidth}{!}{
\begin{tabular}{|l|c|c|c|c|c|}
\hline
& SpatialGAN \cite{SpatialGAN} & PS-GAN \cite{PS_GAN} & $\infty$-GAN \cite{InfinityGAN} & LocoGAN \cite{LocoGAN}  & LocoGAN~\cite{LocoGAN}+SG2~\cite{LocoGAN}+Fourier~\cite{SIREN,FourierINR}\\
\hline
Has local latents? & \checkmark & \checkmark & \checkmark & \checkmark & \checkmark  \\
Has global latents? & \checkmark & \checkmark & \checkmark & \checkmark & \checkmark  \\
Has periodic positional embeddings? &  & \checkmark & \checkmark & \checkmark & \checkmark  \\
Not limited to a single image? & \checkmark & \checkmark & & \checkmark & \checkmark  \\
Was employed for non-texture datasets? &  &  & N/A & \checkmark & \checkmark  \\
Is big? &  &  &  & & \checkmark \\
\hline
\end{tabular}
}
\end{table}

\Q{How is your method different from panoramas generation?}

\A A panorama is produced by panning, i.e. camera rotation, while our method performs \textit{tracking}, i.e. camera transition.
Also, a (classical) panorama is limited to $360^\circ$ and will start repeating itself in the same manner as LocoGAN does in Figure~\ref{fig:qualitative-comparison}.

\Q{How is your method different from image stitching?}

\A Image stitching connects nearby images of the \textit{same} scene into a single large image, while our method merges \textit{entirely} different scenes.

\Q{Does your framework work for 2D (i.e., joint vertical + horizontal) infinite generation? Why didn't you explore it in the paper?}

\A Yes, it can be easily generalized to this setup by positioning anchors not on a 1D line (as shown in Figure~\ref{fig:alignment}), but on a 2D grid instead and using bilinear interpolation instead of the linear one.
The problem with 2D infinite generation is that there are no \textit{complex} datasets for this, only pattern-like ones like textures or satellite imagery, where the global context is \textit{unlikely} to span several frames.
And the existing methods (e.g., LocoGAN \cite{LocoGAN}, $\infty$-GAN \cite{InfinityGAN}, PS-GAN \cite{PS_GAN}, SpatialGAN \cite{SpatialGAN}, TileGAN \cite{TileGAN}, etc.) already mastered this setup to a good extent.

\Q{How do you perform instance normalization for layers with $16\times 16$ resolution? Your patch size would be equal to $1\times 1$, so $\bm\sigma (\bm x) = \text{NaN}$ (or $0$) for it.}

\A In contrast to CocoGAN\cite{CocoGAN}, we use \textit{vertical} patches (see Appendix~\apref{ap:implementation-details}), so the minimal size of our patch is $16 \times 1$, hence the std is defined properly.



\Q{Why $\infty$-FID for LocoGAN is so high?}

\A LocoGAN generates a repeated scene and that's why it gets penalized for mode collapse.

\section{Implementation details}\label{ap:implementation-details}


\subsection{Architecture and training}
We preprocess all the datasets with the procedure described in Algorithm~\ref{alg:spatial-inv} with a threshold parameter of 0.7.
We train all the methods on identical datasets to make the comparison fair.

As being said in the main paper, we develop our approach on top of the StyleGAN2 model.
For this, we took the official StyleGAN2-ADA implementation and disabled differentiable augmentations \cite{SGA, DiffAug} since Taming Transformers \cite{TamingTransformers} do not employ such kind of augmentations, so it would make the comparison unfair.
We used a small version of the model (\textsf{config-e} from \cite{StyleGAN2}) for our $256^2$ experiments and the large one (\textsf{config-f} from \cite{StyleGAN2}) for $1024^2$ experiments.
They differ in the number of channels in high-resolution layers: the \textsf{config-f}-model uses twice as big dimensionalities for them.

We preserve all the training hyperparameters the same.
Specifically, we use $\gamma = 10$ for R1-regularization \cite{R1_reg} for all StyleGAN2-based experiments.
We use the same probability of 0.9 for style mixing.
In our case, since on each iteration we input 3 noise vectors $\bm w_l, \bm w_c, \bm w_r$ we perform style mixing on them independently.
Following StyleGAN2, we also apply the Perceptual Path Loss \cite{StyleGAN2} regularization with the weight of 2 for our generator $\G$.
We also use the Adam optimizer to train the modules with the learning rates of 0.0025 for $\G$ and $\D$ and betas of 0.0 and 0.99.

\begin{figure}
    \centering
    \begin{subfigure}[b]{0.49\linewidth}
        \centering
        \includegraphics[width=0.8\textwidth]{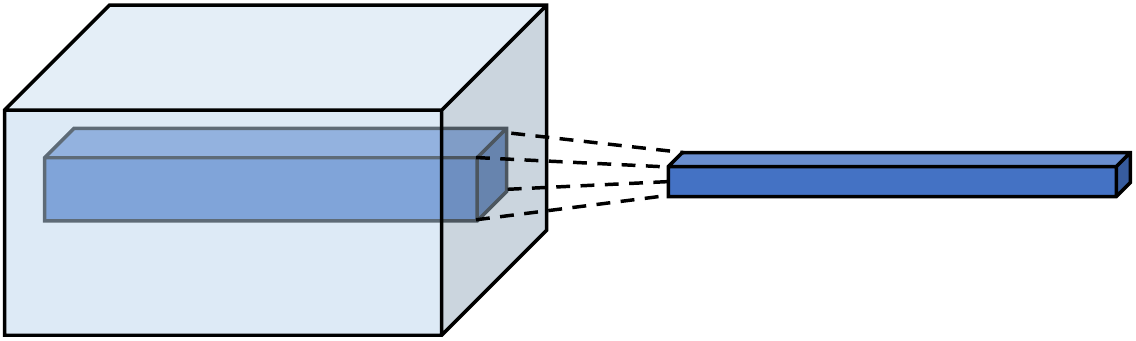}
        \caption{\textsf{Conv3x3}}
        \label{fig:coordconv:conv}
    \end{subfigure}
    \hfill
    \begin{subfigure}[b]{0.49\linewidth}
        \centering
        \includegraphics[width=0.8\textwidth]{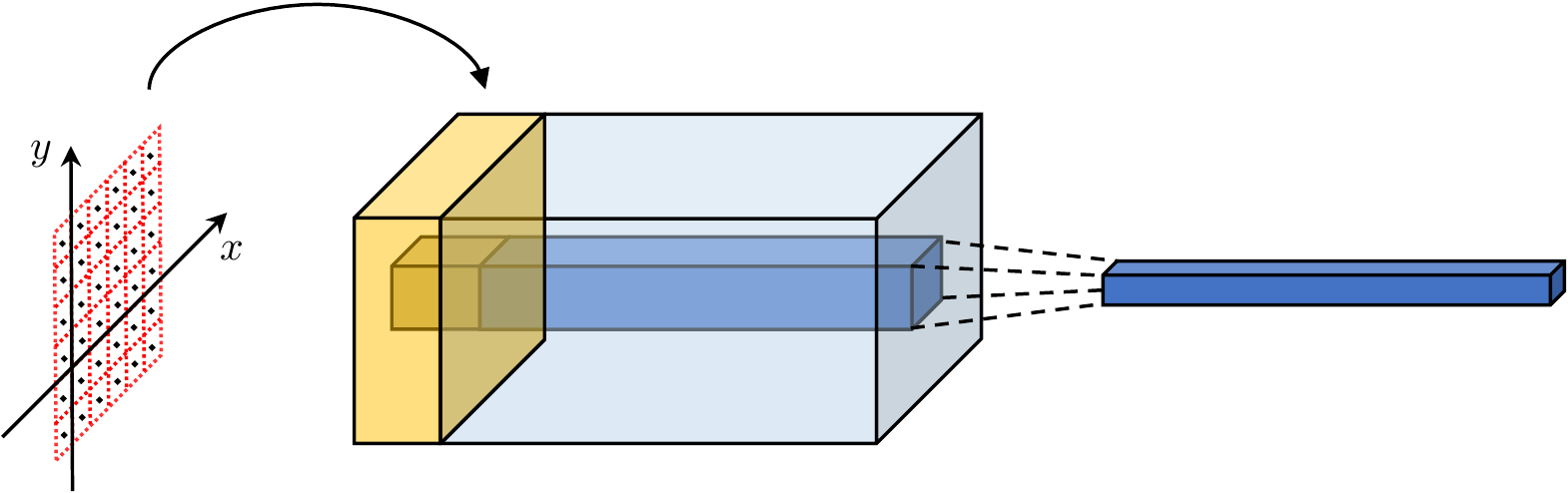}
        \caption{\textsf{CoordConv3x3}. Orange part of the hidden representation denotes coordinates embeddings.}
        \label{fig:coordconv:coordconv}
    \end{subfigure}
    \caption{Illustration of \textsf{Conv3x3} and \textsf{CoordConv3x3}. For \textsf{CoordConv3x3} there are different strategies on how to compute coordinates embeddings, for ALIS, we use periodic positional encoding \cite{SIREN, FourierINR}.}
    \label{fig:coordconv}
\end{figure}

As being said in the main paper, \textsf{CoordConst} and \textsf{CoordConv3x3} are analogs of StyleGAN's \textsf{Const} block and \textsf{Conv3x3} respectively, but with coordinates positional embeddings concatenated to the hidden representations.
We illustrate this in Figure~\ref{fig:coordconv}.
As being said in Section~\ref{sec:method}, our \textsf{Conv3x3} is not modulated because one cannot make the convolutional weights be conditioned on a position efficiently and we use SA-AdaIN instead to input the latent code.
\textsf{Const} block of StyleGAN2 is just a learnable 3D tensor of size $512\times 4 \times 4$ which is passed as the starting input to the convolution decoder.
For \textsf{CoordConst} layer, we used padding mode of \textsf{repeat} for its constant part to make it periodic and not to break the equivariance.

To avoid numerical issues, we sampled $\delta$ only at discrete values of the interval $[0, 2d - W]$, corresponding to a left border of each patch.

Each StyleGAN2-based model was trained for 2500 kimgs (i.e. until $\D$ saw 25M real images) with a batch size of 64 (distributed with individual batch size of 16 on 4 GPUs).
Training cumulatively took 2.5 days on 4 V100 GPUs.

For Taming Transformer, we trained the model for 5 days in total: 2.5 days on 4 V100 GPUs for the VQGAN component and 2.5 days on 4 V100 GPUs for the Transformer part.
VQGAN was trained with the cumulative batch size of 12 and the Transformer part --- with 8.
But note that in total it was consuming \textit{more} GPU memory compared to our model: 10GB vs 6GB per a GPU card.
We used the official implementation with the official hyperparameters\footnote{\href{https://github.com/CompVis/taming-transformers}{https://github.com/CompVis/taming-transformers}}.

\begin{figure*}
    \centering
    \includegraphics[width=0.6\textwidth]{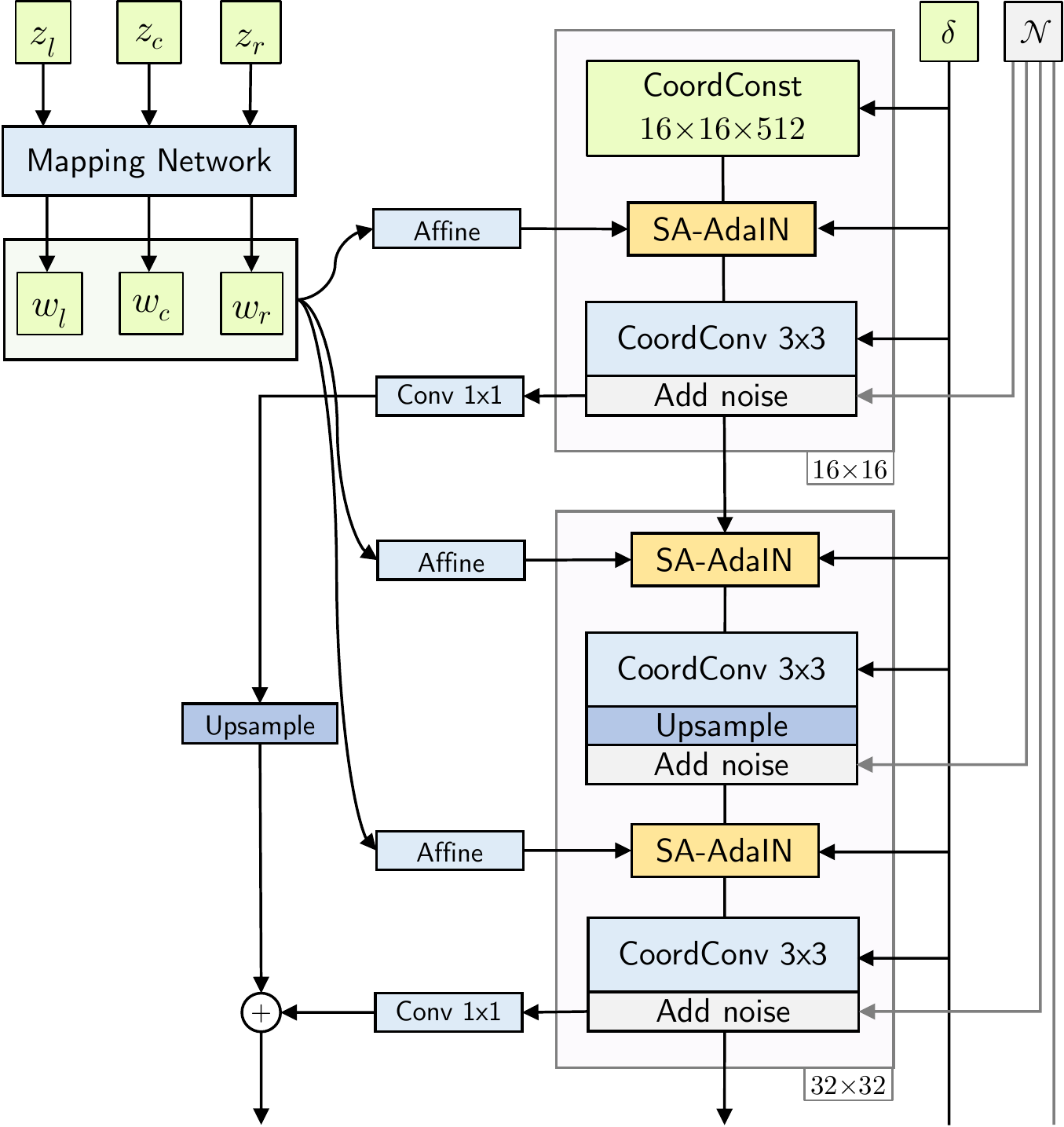}
    \caption{Full architecture of our $\G$.}
    \label{fig:architecture-full}
\end{figure*}

\begin{figure}
    \centering
    \begin{subfigure}[b]{0.33\linewidth}
        \centering
        \includegraphics[width=0.8\textwidth]{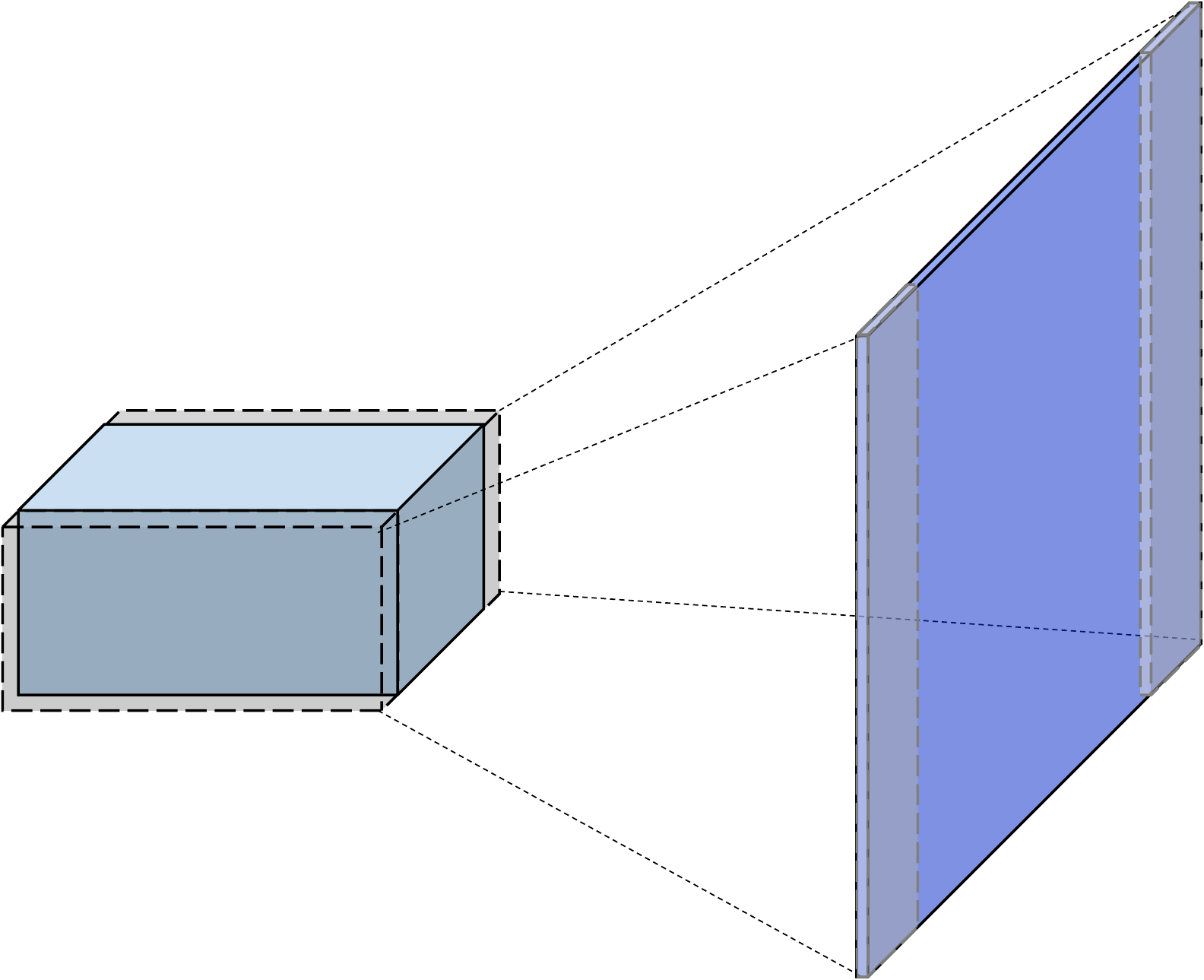}
        \caption{Padding problem}
        \label{fig:inf-gen-strategies:padding-problem}
    \end{subfigure}
    \hfill
    \begin{subfigure}[b]{0.33\linewidth}
        \centering
        \vspace{0.5cm}
        \includegraphics[width=0.8\textwidth]{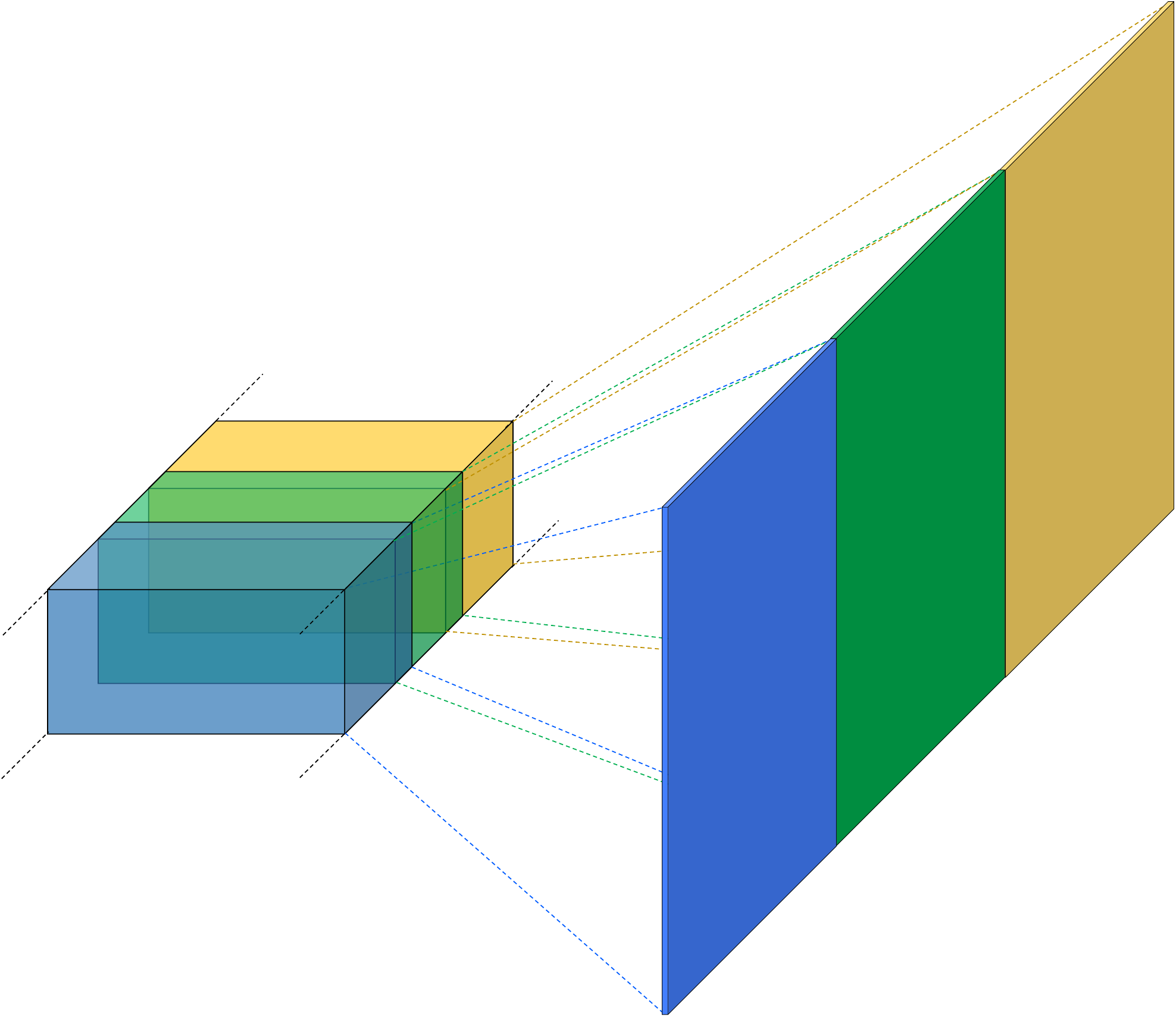}
        \caption{Overlapped generation \cite{SpatialGAN, LocoGAN, PS_GAN, InfinityGAN, TamingTransformers}}
        \label{fig:inf-gen-strategies:overlapped}
    \end{subfigure}
    \hfill
    \begin{subfigure}[b]{0.33\linewidth}
        \centering
        \vspace{0.5cm}
        \includegraphics[width=0.8\textwidth]{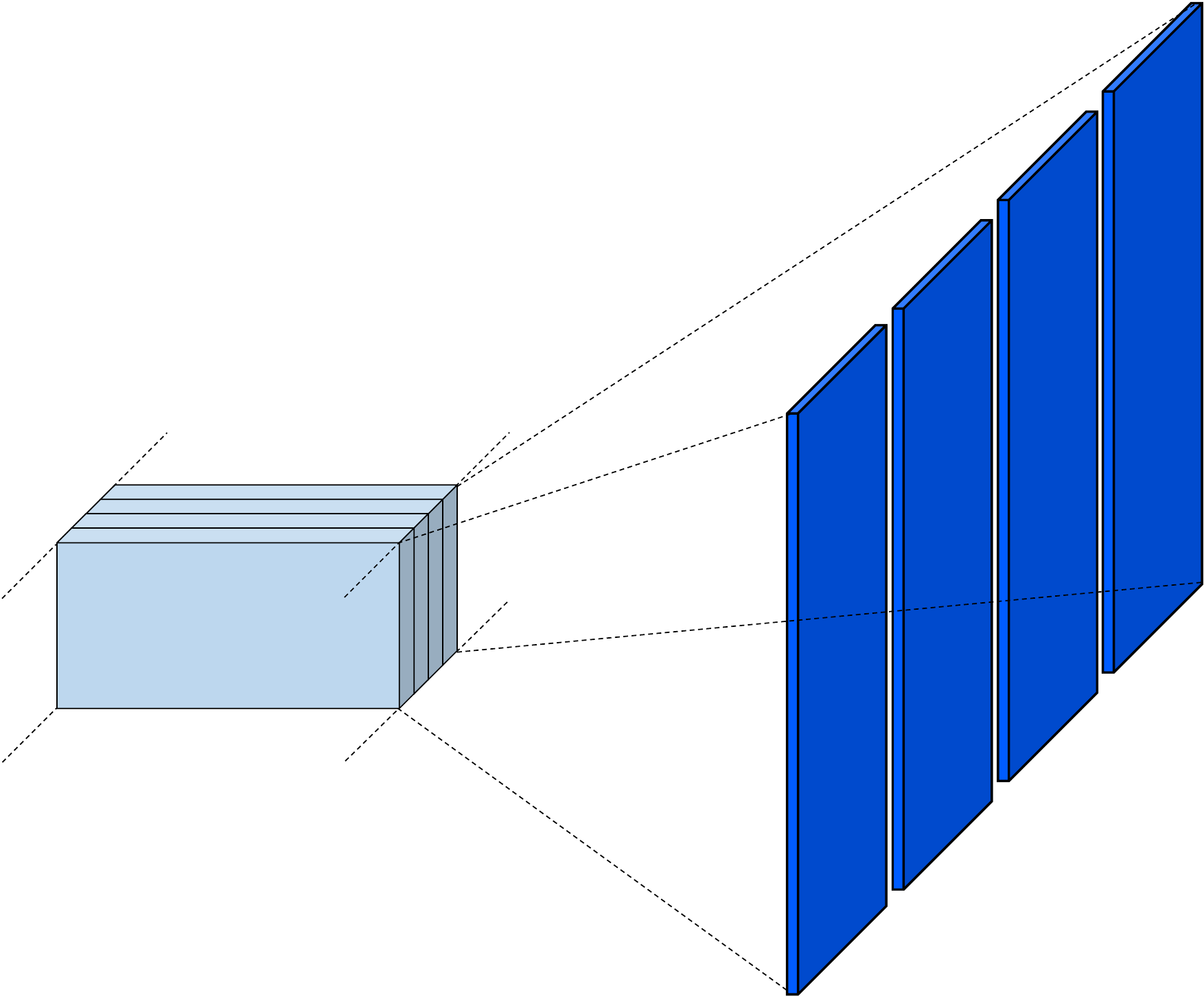}
        \caption{Patchwise generation \cite{CocoGAN, INR_GAN, CIPS}}
        \label{fig:inf-gen-strategies:patchwise}
    \end{subfigure}
    \caption{(a) A padding problem that occurs with infinite image generation; (b, c) two strategies to alleviate it. For ALIS, we use patchwise generation and show how to incorporate spatially varying \textit{global} latent codes into it.}
    \label{fig:inf-gen-strategies}
\end{figure}

\subsection{Patchwise generation}
As being said in Section~\ref{sec:method}, following the recent advances in coordinate-based generators \cite{CocoGAN, CIPS, INR_GAN}, our method generates images via spatially independent patches, which is illustrated in Figure~\ref{fig:inf-gen-strategies:patchwise}.
The motivation of it is to force the generator learn how to stitch nearby patches based on their positional information and latent codes.
This solved the ``padding'' problem of infinite image generation: since traditional decoders extensively use padding in their implementations, this creates stitching artifacts when the produced frames are merged naively, because padded values are actually filled with neighboring context during the frame-by-frame infinite generation (see Figure~\ref{fig:inf-gen-strategies:padding-problem}).
A traditional way to solve this \cite{SpatialGAN, PS_GAN, LocoGAN, InfinityGAN} is to merge the produced frames with overlaps, like depicted in Figure~\ref{fig:inf-gen-strategies:overlapped}.

\section{Which datasets are connectable?}\label{ap:spatial-inv}

\begin{figure}
    \centering
    \includegraphics[width=0.5\textwidth]{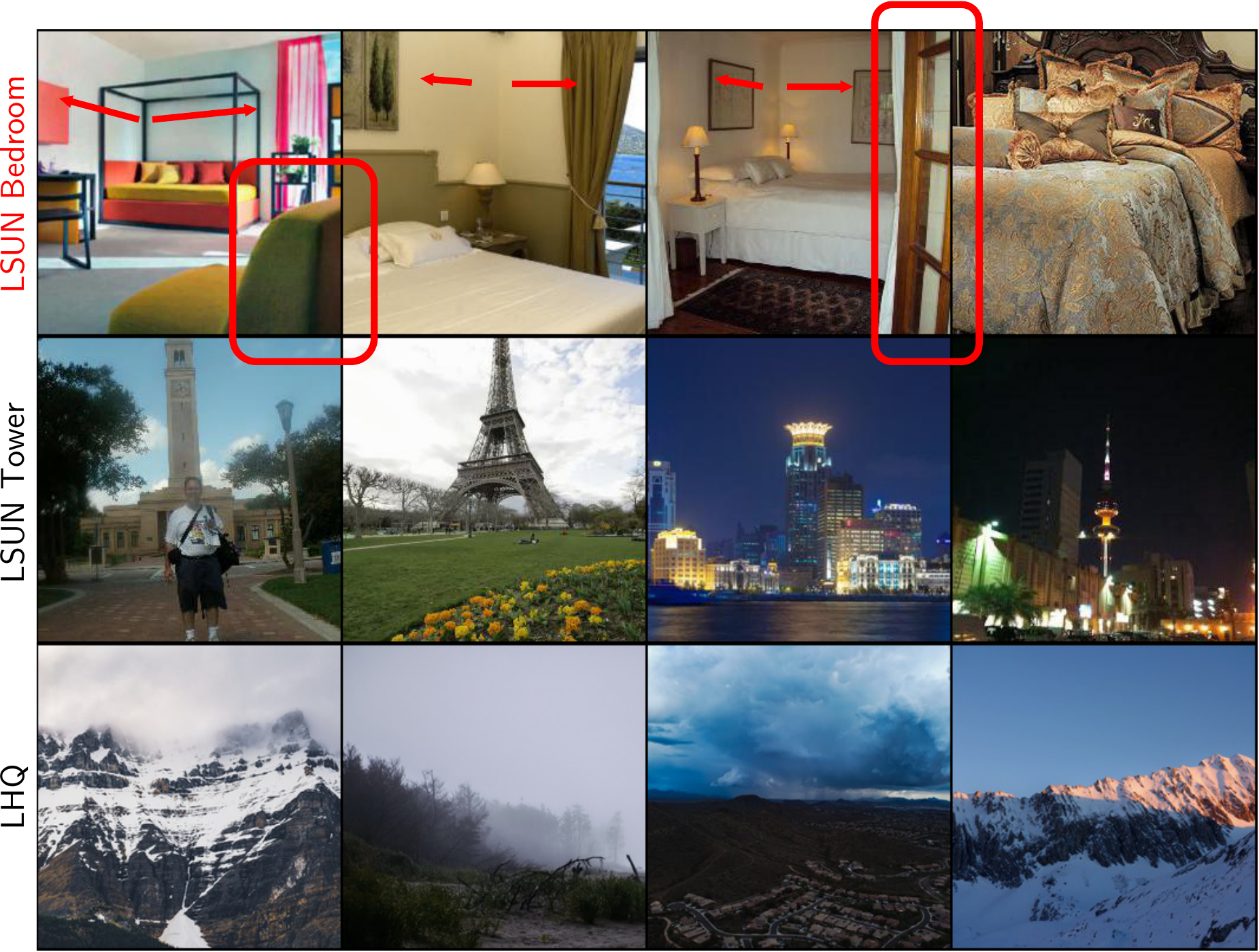}
    \caption{Random samples from LSUN Bedroom (top), LSUN Tower (middle) and LHQ (bottom). While LSUN Tower and LHQ have (approximately) horizontally invariant statistics, for LSUN Bedroom this does not hold, which is highlighted in red. For LSUN Bedroom, the first problem occurs due to walls: it is impossible to continue a sample to the left due to wall hitting the camera. And there are no samples in LSUN Bedroom which would have the transition between those walls directions. The second problem is in too-close-to-camera objects. We propose Algorithm \ref{alg:spatial-inv} to test if the dataset is connectable and to find a connectable subset of it. After preprocessing LSUN Bedroom, our algorithm can fit it, as shown in Figure~\ref{fig:connecting-lsun-bedroom}. The importance of this procedure is also confirmed by FID scores in Table~\ref{table:conf-thresholds}.}
    \label{fig:spatial-inv-problems}
\end{figure}

In this section, we elaborate on the importance of having spatially invariant statistics for training an infinite image generator.
Since we consider only horizontal infinite image generation, we are concerned about horizontally invariant statistics, but the idea can be easily generalized to vertical+horizontal generation.
As noted by \cite{TamingTransformers}, to train a successful ``any-aspect-ratio'' image generator from a dataset of just independent frames, there should be images containing not only scenes, but also transitions between scenes.
We illustrate the issue in Figure~\ref{fig:spatial-inv-problems}.
And this property does not hold for many computer vision datasets, for example FFHQ, CelebA, ImageNet, LSUN Bedroom, etc.
To test if a dataset contains such images or not and to extract a subset of good images, we propose the following procedure.

Given a dataset, train a binary classifier to predict whether a given half of an image is the left one or right one.
Then, if the classifier has low confidence on the test set, the dataset contains images with spatially invariant statistics, since image left/right patch locations cannot be confidently predicted.
Now, to extract the subset of such good images, we just measure the confidence of a classifier and pick images with low confidence, as described in Algorithm~\ref{alg:spatial-inv}.

To highlight the importance of this procedure, we trained our model on different subsets of LHQ and LSUN Tower, varying the confidence threshold $t$.
As one can see from Table~\ref{table:conf-thresholds}, our procedure is essential to obtain decent results.
Also the confidence threshold should not be too small, because too few images are left making the training unstable.

As a classifier, we select an ImageNet-pretrained WideResnet50 and train it for 10 epochs with Adam optimizer and learning rates of 1e-5 for the body and 1e-4 for the head.
We consider only square center crops of each image.

Also, we tested our ``horizontal invariance score'' for different popular datasets and provide them in Table~\ref{table:spatial-inv-scores}.
As one can see from this table, our proposed dataset contains images with much more spatially invariant statistics.

We used $t=0.95$ for LHQ and $t=0.7$ for LSUN datasets.

\begin{algorithm}
\normalsize
\SetAlgoLined
\SetKwInOut{Input}{Input}\SetKwInOut{Output}{Output}
\Input{Dataset $D$ of $(H \times W \times C)$-sized images}
\Input{Confidence threshold $t$.}
\Output{Dataset $D_t \subseteq D$ of horizontally invariant images.}

Initialize a binary classifier $C_\theta$\;
Split $D$ into subsets $D_\text{train}, D_\text{val}, D_\text{rest}$ with ratio 2:1:7\;

\While{not converged}{
    Sample a training batch $\bm X \subseteq D_\text{train}$\;
    \For{$i\leftarrow 1$ \KwTo $|\bm X|$}{
        Randomly sample side $s^{(i)} \in \{\text{left}, \text{right}\}$\;
        \eIf{$s^{(i)}$ = \textnormal{left}}{
            $x^{(i)} \leftarrow x^{(i)}[:, :W/2]$ \tcp*[l]{Select the left half}
            $y^{(i)} \leftarrow 0$\;
        }{
            $x^{(i)} \leftarrow x^{(i)}[:, W/2:]$ \tcp*[l]{Select the right half}
            $y^{(i)} \leftarrow 1$\;
        }
  }
  Update $C_\theta$ based on training batch $\{ x^{(i)}, y^{(i)} \}_{i=1}^{|\bm X|}$\;
}
Evaluate $C_\theta$ on $D_\text{rest}$ to obtain scores for each image half $\bm R = (r^{(1)}_\text{left}, r^{(1)}_\text{right}, r^{(2)}_\text{left}, ..., r^{(N)}_\text{left}, r^{(N)}_\text{right})$\;
Set $D_t \leftarrow \{x^{(i)}| x^{(i)} \in D_\text{rest} \wedge \max \{ r^{(i)}_\text{left}, r^{(i)}_\text{right} \} < t \}$\;
Return $D_t$\;
\caption{Extract a subset with (approximately) horizontally invariant statistics.}
\label{alg:spatial-inv}
\end{algorithm}

\begin{table*}
\caption{FID scores of ALIS for different confidence thresholds $t$ in Algorithm~\ref{alg:spatial-inv} on LHQ $256^2$ and LSUN Tower $256^2$. Preprocessing the dataset with the proposed procedure is essential to obtain decent results, since it filters out ``bad'' images, i.e. images with spatially non-invariant statistics. We select $t=0.7$ for LSUN datasets and $t=0.95$ for LHQ since decreasing $t$ filters away too many images which also hurts the FID score. ``Dataset size'' is the number of images remaining after the filtering procedure. For $t=1$, the dataset is untouched.}
\label{table:conf-thresholds}
\centering
\begin{tabular}{|l|cc|cc|}
\hline
\multirow{2}{*}{Method} & \multicolumn{2}{c|}{LSUN Tower $256^2$} & \multicolumn{2}{c|}{LHQ $256^2$} \\
& FID & Dataset size & FID & Dataset size \\
\hline
$t=1$ & 17.51 & 708k & 23.23 & 90k \\
$t=0.99$ & 8.68 & 168k & 10.18 & 50k \\
$t=0.95$ & 8.85 & 116k & 10.48 & 38k \\
$t=0.7$ & 8.83 & 59k & 11.49 & 19k \\
$t=0.5$ & 9.73 & 41.5k & 14.55 & 13k \\
\hline
\end{tabular}
\end{table*}

\begin{table}
\caption{Spatial invariance scores for different datasets. We used Algorithm \ref{alg:spatial-inv} to train a classifier and computed its mean accuracy on $D_\text{rest}$. In all the cases, we used center crops.}
\label{table:spatial-inv-scores}
\centering
\begin{tabular}{|l|c|}
\hline
Dataset & Mean Test Confidence \\
\hline
LSUN Bedroom & 95.2\% \\
LSUN Tower & 92.7\% \\
LSUN Bridge & 88.6\% \\
LSUN Church & 96.1\% \\
FFHQ & 99.9\% \\
ImageNet & 99.8\% \\
\hline
LHQ & 82.5\% \\
\hline
\end{tabular}
\end{table}

\section{Landscapes HQ dataset}\label{ap:lhq}

To collect the dataset we used two sources: Unsplash\footnote{\href{https://unsplash.com/}{https://unsplash.com/}} and Flickr\footnote{\href{https://www.flickr.com/}{https://www.flickr.com/}}.
For the both sources, the acquisition procedure consisted on 3 stages: constructing search queries, downloading the data, refining the search queries based on manually inspecting the subsets corresponding to different search queries and manually constructed blacklist of keywords.

After a more-or-less refined dataset collection was obtained, a pretrained Mask R-CNN model was employed to filter out the pictures which likely contained an object.
The confidence threshold for ``objectivity'' was set to 0.2.

For Unsplash, we used their publicly released dataset of photos descriptions\footnote{\href{https://unsplash.com/data}{https://unsplash.com/data}} to extract the download urls.
It was preprocessed by first searching the appropriate keywords with a whitelist of keywords and then filtering out images with a blacklist of keywords.
A whitelist of keywords was constructed the following way.
First, we collected 230 prefixes, consisting on geographic and nature-related places.
Then, we collected 130 nature-related and object-related suffixes.
A whitelist keyword was constructed as a concatenate of a prefix and a suffix, i.e. $230 \times 130 = 30k$ whitelist keywords.
Then, we extracted the images and for each image enumerated all its keywords and if there was a keyword from our blacklist, then the image was removed.
A blacklist of keywords was constructed by trial and error by progressively adding the keywords from the following categories: animals, humans, human body parts, human activities, eating, cities, buildings, country-specific sights, plants, drones images (since they do not fit our task), human belongings, improper camera positions.
In total, it included 300 keywords.
We attach both the whitelist and the blacklist that were used in the supplementary.
In total, 230k images was collected and downloaded with this method.

For Flickr, we constructed a set of 80 whitelist keywords and downloaded images using its official API\footnote{\href{https://www.flickr.com/services/api/}{https://www.flickr.com/services/api/}}.
They were much more strict, because there is no way to use a blacklist.
In total, 170k images were collected with it.

After Unsplash and Flickr images were downloaded, we ran a pretrained Mask R-CNN model and removed those photos, for which objectivity confidence score was higher than 0.2.
In total, that left 60k images for Unsplash and 30k images for Flickr.

The images come in one of the following licenses: Unsplash License, Creative Commons BY 2.0, Creative Commons BY-NC 2.0, Public Domain Mark 1.0, Public Domain CC0 1.0, or U.S. Government Works license.
All these licenes allow the use of the dataset for research purposes.

\begin{figure*}
    \centering
    \includegraphics[width=\textwidth]{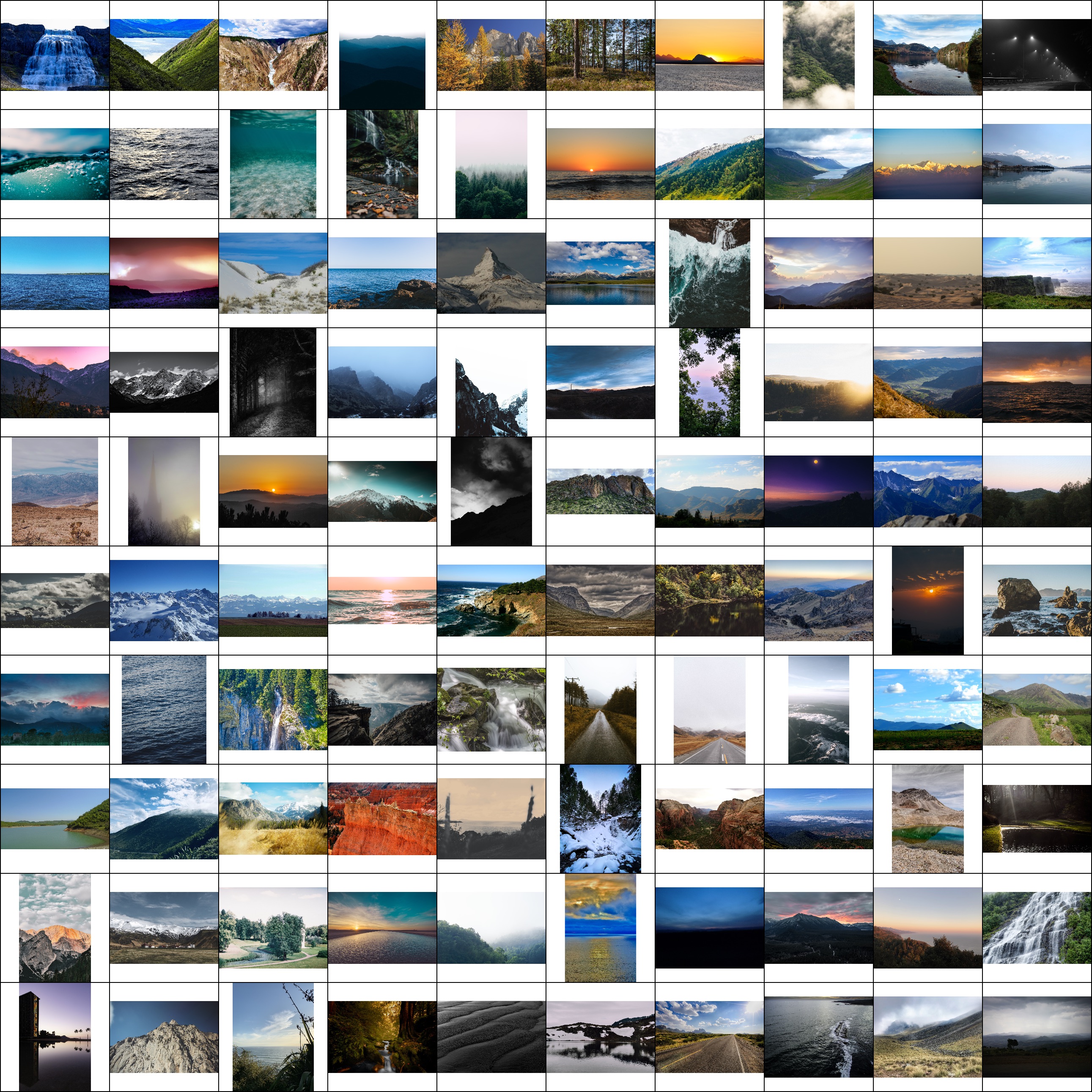}
    \caption{100 random images from LHQ dataset, that we introduce in our work. It is a diverse dataset of very different scenes and in high resolution $\geq 1024^2$. We downsized the images for this figure to avoid performance issues.}
    \label{fig:lhq-random-images}
\end{figure*}
\begin{figure}
    \centering
    \includegraphics[width=0.8\linewidth]{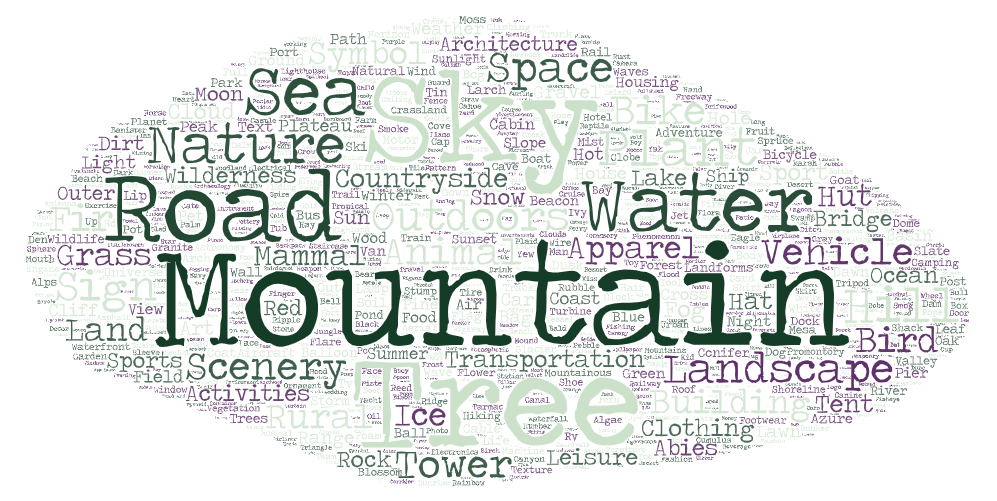}
    \caption{A word map of keywords in the LHQ dataset.}
    \label{fig:keywords-word-map}
\end{figure}

We depict 100 random samples from LHQ in Figure~\ref{fig:lhq-random-images} and a word map of keywords in Figure~\ref{fig:keywords-word-map}.

\section{Remarks on comparison to Taming Transformers}\label{ap:taming-transformers}

\begin{figure}
    \centering
    \includegraphics[width=\textwidth]{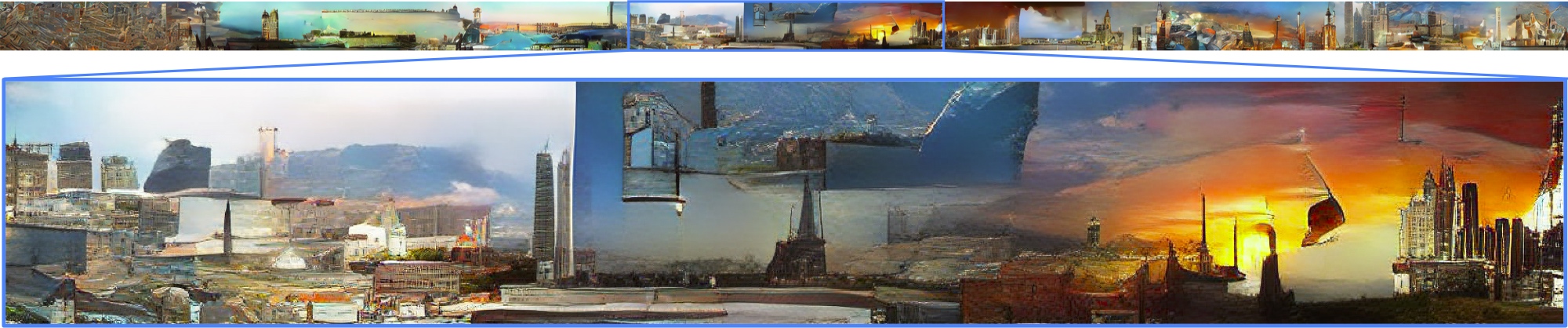}
    \\
    \vspace{1em}
    \includegraphics[width=\textwidth]{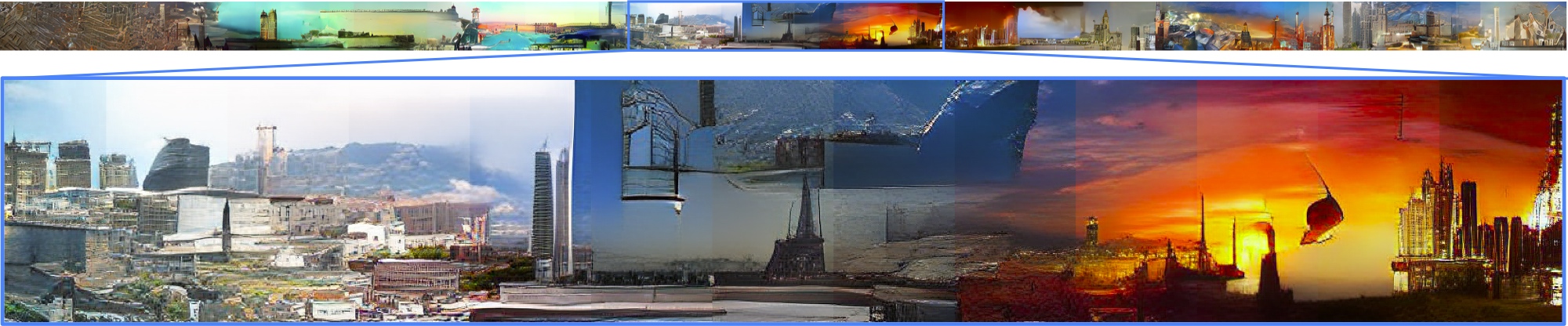}
    \caption{Illustration of GroupNorm artifacts of TT \cite{TamingTransformers} when frame-by-frame infinite generation is considered. Top: normal samples, bottom: overlapped generation (see Figure~\ref{fig:inf-gen-strategies}). Since TT's decoder uses GroupNorm \cite{GroupNorm}, it computes activations statistics across the whole image, i.e. it needs the whole image to be available simultaneously during inference. If frame-by-frame generation is employed at test-time (to produce an infinite image, for example), then it leads to the illustrated artifacts.}
    \label{fig:tt-artifacts}
\end{figure}

Using Taming Transformers (TT) \cite{TamingTransformers} as a baseline is complicated for the following reasons:
\begin{itemize}
    \item The original paper was mainly focused on conditional image generation. Since our setup is unconditional (and thus harder), the produced samples are of lower quality and it might confuse a reader when comparing them to conditionally generated samples from the original TT's paper \cite{TamingTransformers}.
    \item It is not a GAN-based model (though there is some auxiliary adversarial loss added during training) --- thus, in contrast to LocoGAN or ALIS, it cannot be reimplemented in the StyleGAN2's framework. This makes the comparison to StyleGAN2-based models harder since StyleGAN2 is so well-tuned ``out-of-the-box''. However, it must be noted that TT is a very recent method and thus was developed with access to all the recent advances in generative modeling. As being said in Section~\ref{sec:experiments}, we used the official implementation with the official hyperparameters for unconditional generation training. In total, it was trained for twice as long compared to the rest of the models: 2.5 days on 4 V100 GPUs for the VQGAN part and 2.5 days on 4 V100 GPUs for the Transformer part.
    \item Autoregressivy inference makes it very slow at test time. For example, it took 8 days to compute FID and $\infty$-FID scores on a single V100 GPU for a single experiment.
    \item Taming Transformer's decoder uses GroupNorm layer \cite{GroupNorm}. GroupNorm, as opposed to other normalization layers (like BatchNorm \cite{BatchNorm}) does not collect running statistics and computes from the hidden representation even at test-time. It becomes problematic when generating an ``infinite'' image frame by frame, because neighboring frames use different statistics at forward pass. This is illustrated in Figure~\ref{fig:tt-artifacts}. To solve the for $\infty$-FID, instead of generating a long $256\times(50000 \cdot 256)$ image, we generated 500 $256 \times (100 \cdot 256)$ images. Note that it makes the task easier (and improves the score) because this increases the diversity of samples, which FID is very sensitive to.
\end{itemize}

\section{Additional samples}\label{ap:samples}

\begin{figure}
    \centering
    \includegraphics[width=\textwidth]{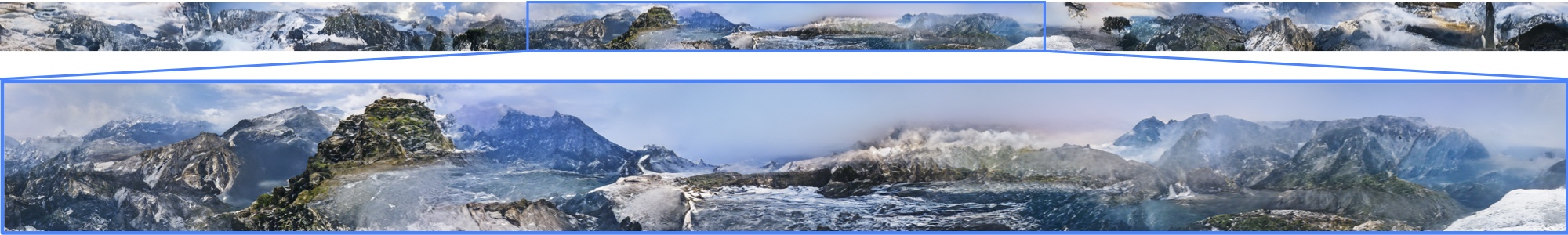}
    \\
    \vspace{1em}
    \includegraphics[width=\textwidth]{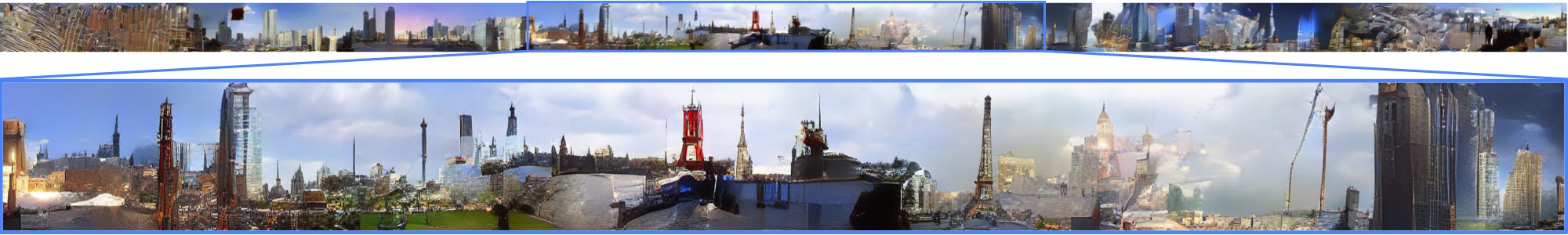}
    \\
    \vspace{1em}
    \includegraphics[width=\textwidth]{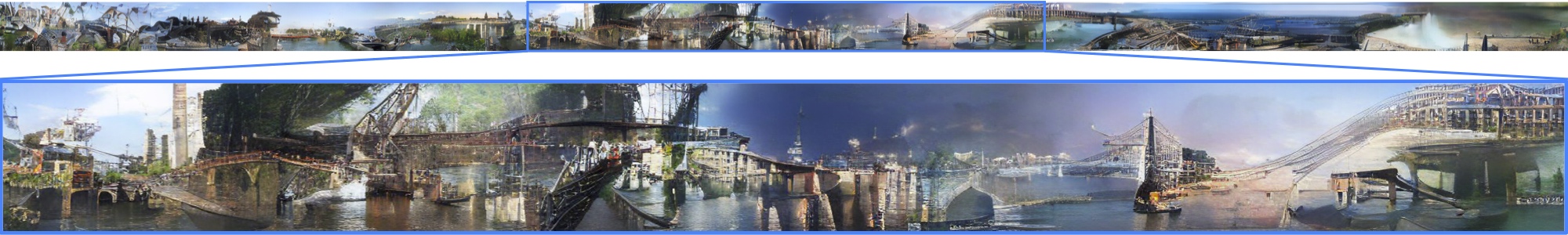}
    \caption{Random samples from Taming Transformer \cite{TamingTransformers} on LHQ, LSUN Tower and LSUN Bridge}
    \label{fig:random-samples-tt}
\end{figure}

\begin{figure}
    \centering
    \includegraphics[width=\textwidth]{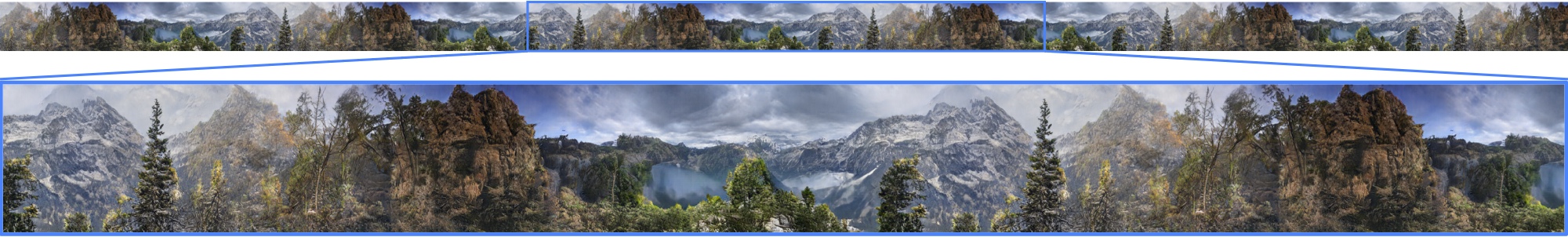}
    \\
    \vspace{1em}
    \includegraphics[width=\textwidth]{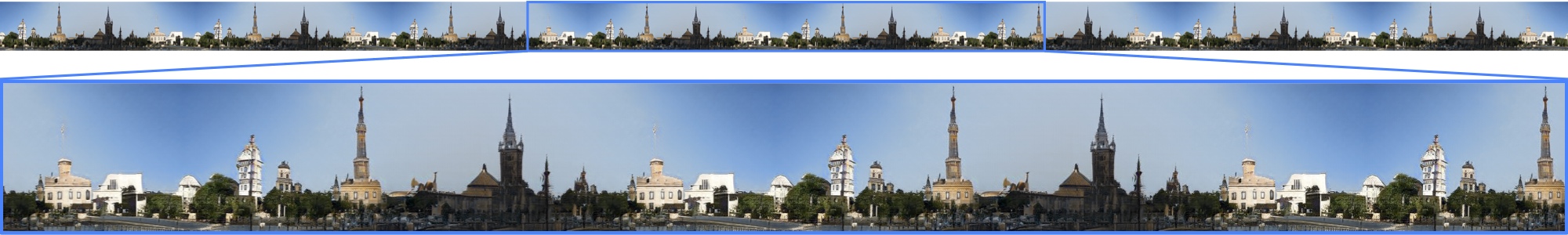}
    \\
    \vspace{1em}
    \includegraphics[width=\textwidth]{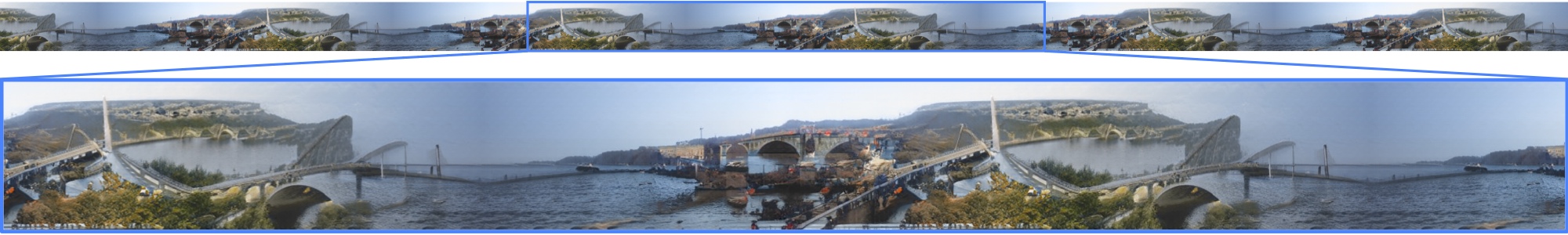}
    \caption{Random samples from LocoGAN+SG2+Fourier on LHQ, LSUN Tower and LSUN Bridge}
    \label{fig:random-samples-locogan}
\end{figure}

\begin{figure}
    \centering
    \includegraphics[width=\textwidth]{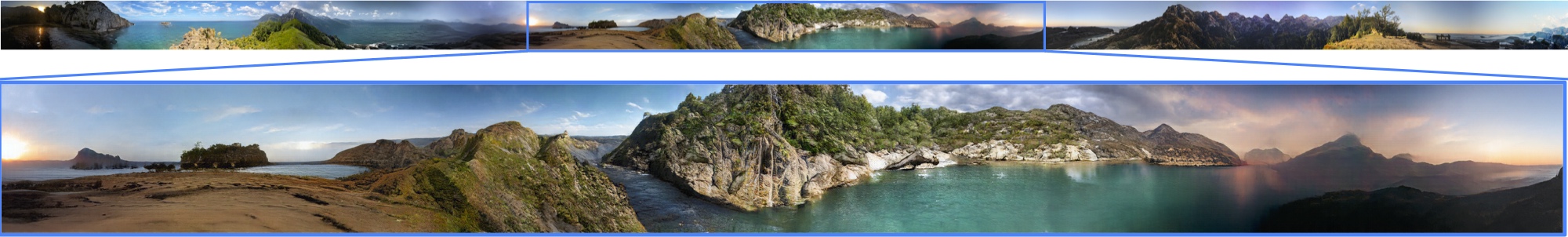}
    \\
    \vspace{1em}
    \includegraphics[width=\textwidth]{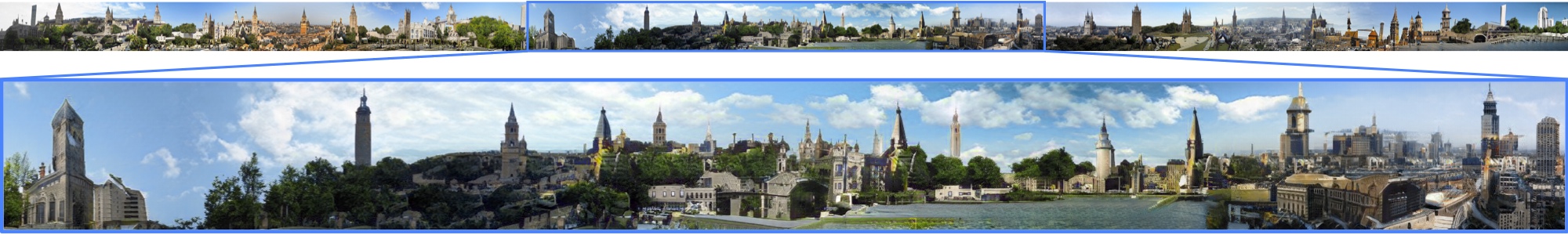}
    \\
    \vspace{1em}
    \includegraphics[width=\textwidth]{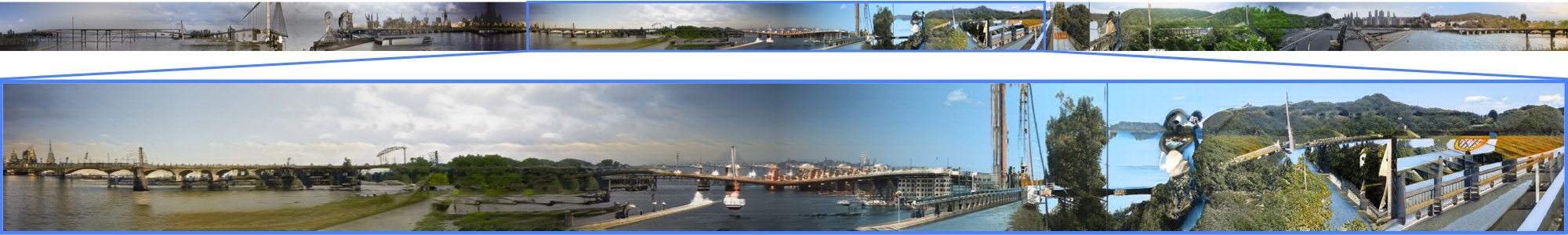}
    \caption{Random samples from ALIS on LHQ, LSUN Tower and LSUN Bridge}
    \label{fig:random-samples-alis}
\end{figure}

\begin{figure}
    \centering
    \includegraphics[width=\textwidth]{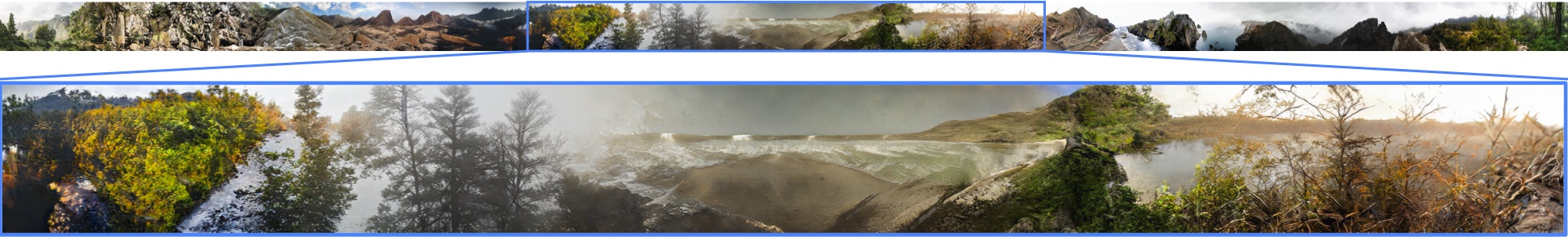}
    \\
    \vspace{1em}
    \includegraphics[width=\textwidth]{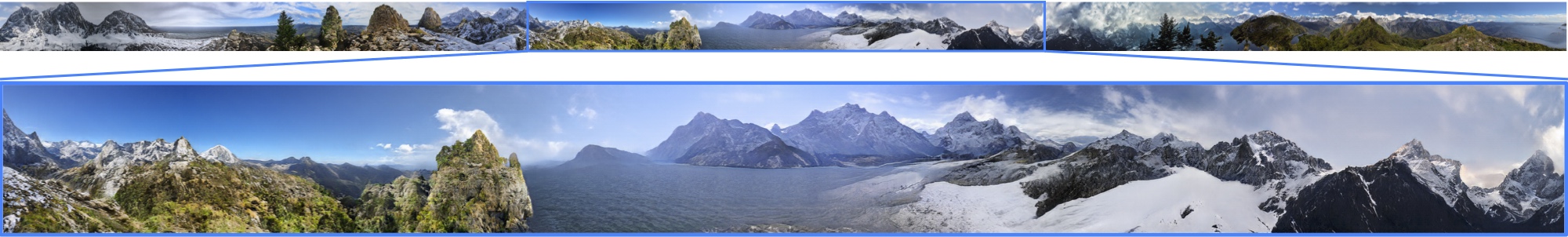}
    \\
    \vspace{1em}
    \includegraphics[width=\textwidth]{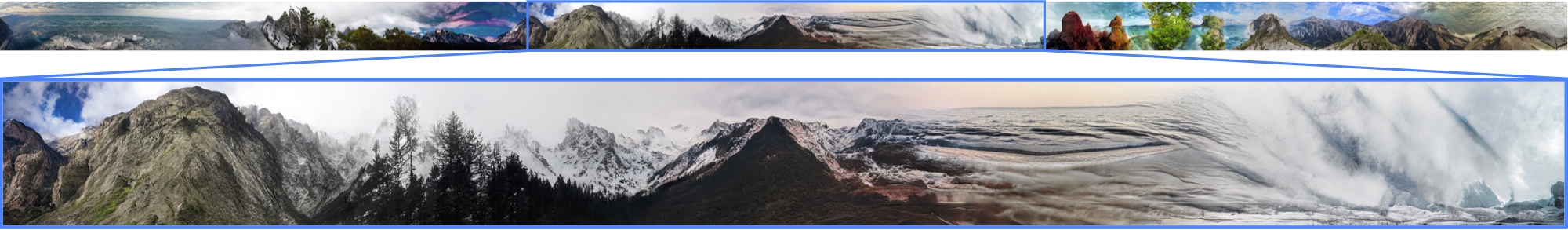}
    \caption{Random samples from ALIS on LHQ. Compare to clustered sampling in Figure~\ref{fig:clustered-sampling}}
    \label{fig:random-samples}
\end{figure}

\begin{figure}
    \centering
    \includegraphics[width=\textwidth]{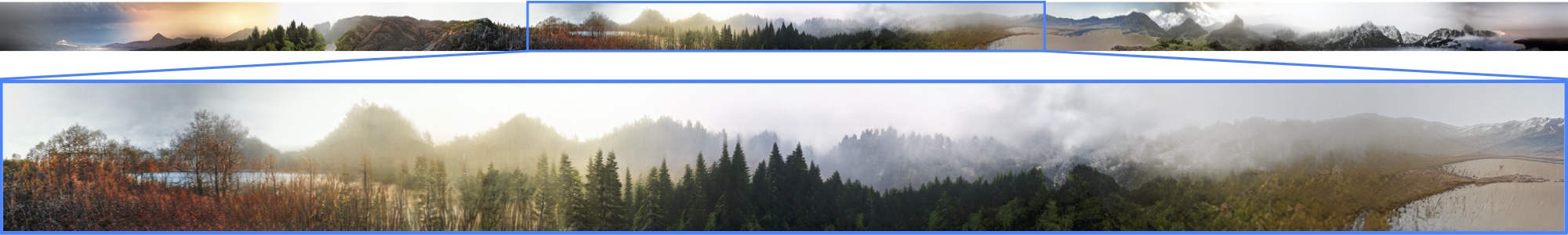}
    \\
    \vspace{1em}
    \includegraphics[width=\textwidth]{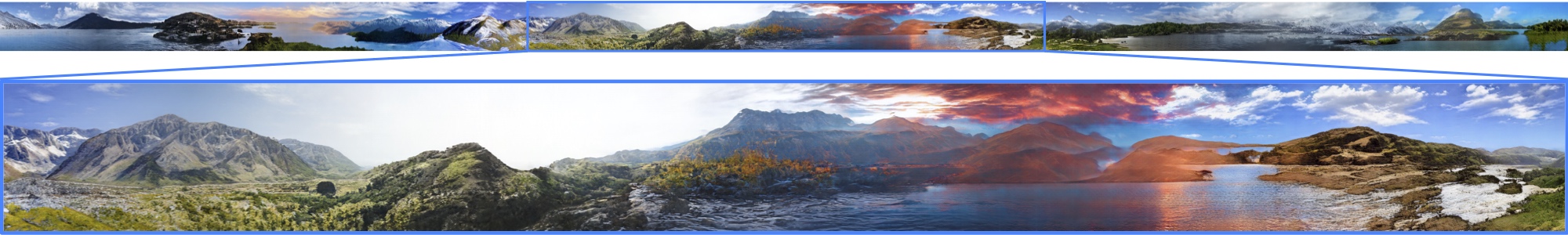}
    \\
    \vspace{1em}
    \includegraphics[width=\textwidth]{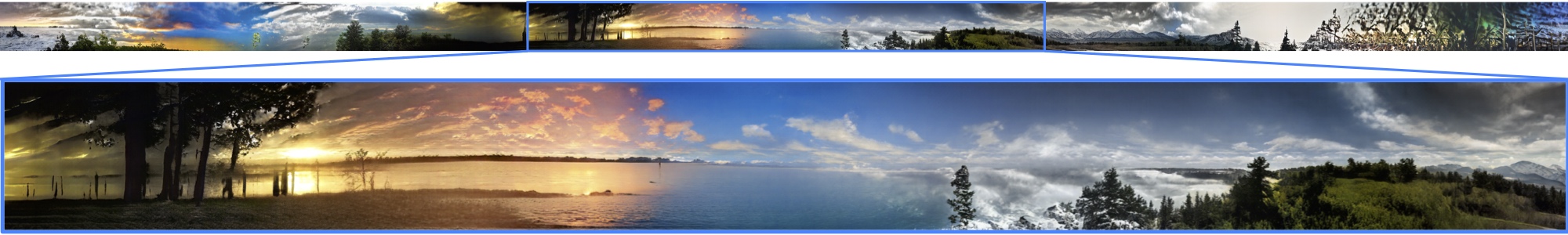}
    \\
    \vspace{1em}
    \includegraphics[width=\textwidth]{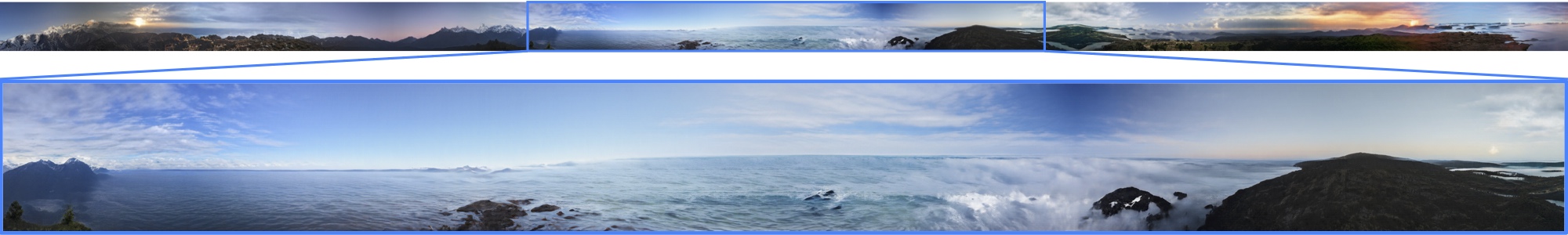}
    \caption{Random samples using ``clustered sampling'' on LHQ. Compare to random samples in Figure~\ref{fig:random-samples}. To generate these images, we trained a Gaussian Mixture Model on 20k latent vectors with 8 components. Then, to generate a single long image, we sampled randomly from only a single mode. In this figure, samples from 4 different modes are presented. This improves sample quality since connecting too different anchors (like close-by water and far-away mountains) produces poor performance (see Figure~\ref{fig:failure-cases}). But this also decreases the diversity of samples.}
    \label{fig:clustered-sampling}
\end{figure}

\begin{figure}
    \centering
    \includegraphics[width=\textwidth]{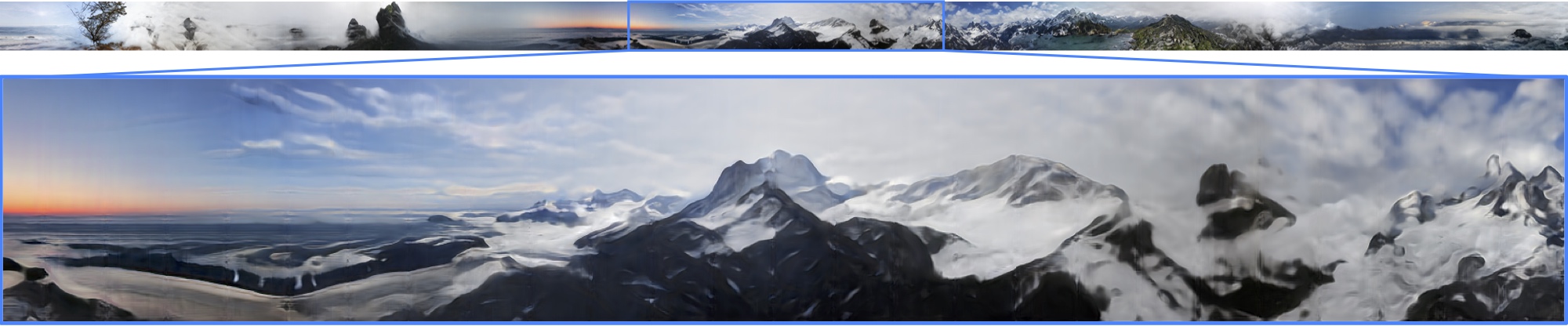}
    \\
    \vspace{1em}
    \includegraphics[width=\textwidth]{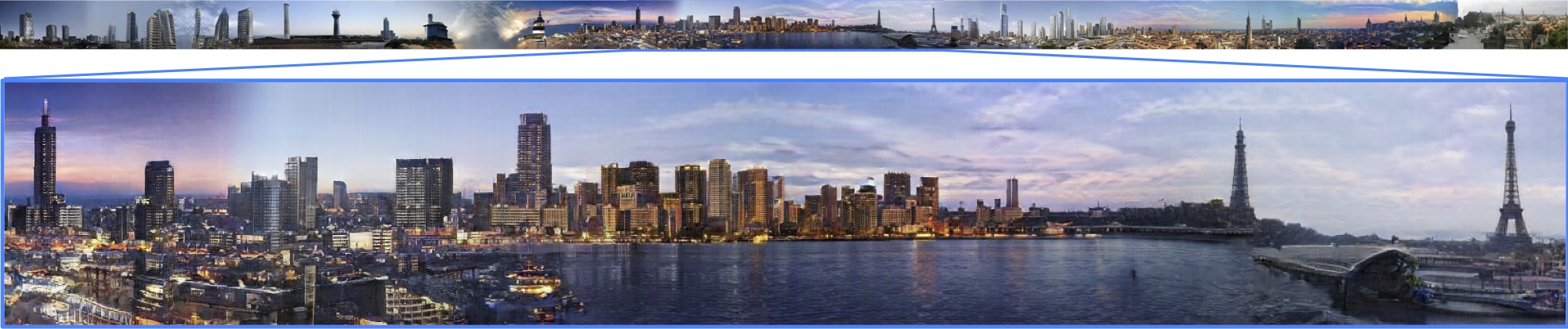}
    \\
    \vspace{1em}
    \includegraphics[width=\textwidth]{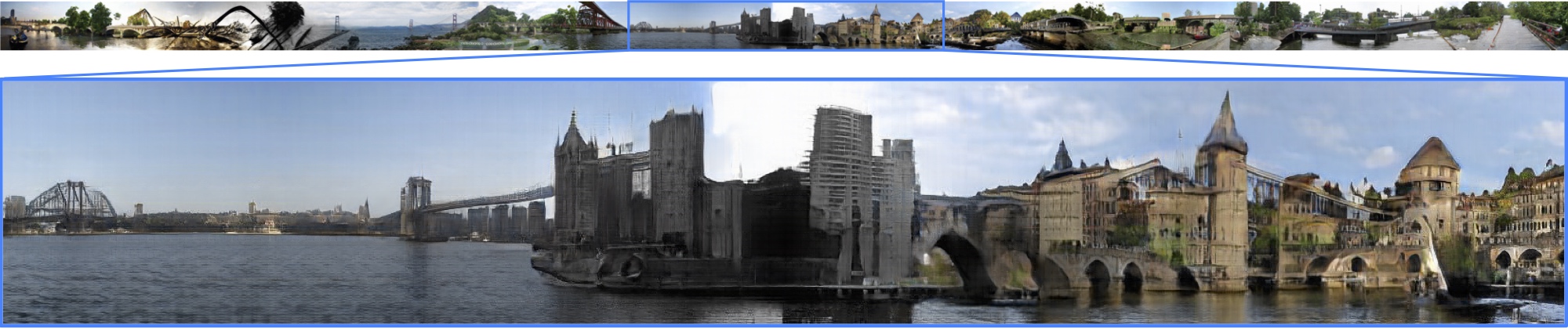}
    \caption{Ablating coordinate embeddings for ALIS on LHQ, LSUN Tower and LSUN Bridge. As being discussed in Section~\ref{sec:experiments}, this leads to blurry generations.}
    \label{fig:random-samples-no-coords}
\end{figure}

\end{document}